\documentclass{article}

\usepackage[ruled,vlined]{algorithm2e}

\usepackage{graphicx}
\usepackage{subfig}

\usepackage{amssymb}
\usepackage{amsmath,bm}
\usepackage{enumitem}
\newtheorem{theorem}{Theorem}

\newtheorem{lemma}{Lemma}

\newtheorem{definition}{Definition}

\newtheorem{proposition}{Proposition}

\usepackage[utf8]{inputenc} 
\usepackage[T1]{fontenc}    
\usepackage{hyperref}       
\usepackage{url}            
\usepackage{booktabs}       
\usepackage{amsfonts}       
\usepackage{nicefrac}       
\usepackage{microtype}      

\usepackage{hyperref}

\usepackage[accepted]{icml2020}

\icmltitlerunning{Neural Bayes}

\begin{document}

\twocolumn[
\icmltitle{Neural Bayes: A Generic Parameterization Method for Unsupervised Representation Learning}




\begin{icmlauthorlist}
\icmlauthor{Devansh Arpit}{sr}
\icmlauthor{Huan Wang}{sr}
\icmlauthor{Caiming Xiong}{sr}
\icmlauthor{Richard Socher}{sr}
\icmlauthor{Yoshua Bengio}{m}
\end{icmlauthorlist}

\icmlaffiliation{sr}{Salesforce Research}
\icmlaffiliation{m}{Mila}

\icmlcorrespondingauthor{Devansh Arpit}{devansharpit@gmail.com}

\icmlkeywords{unsupervised learning}

\vskip 0.3in
]



\printAffiliationsAndNotice{}  

\begin{abstract}
We introduce a parameterization method called Neural Bayes which allows computing statistical quantities that are in general difficult to compute and opens avenues for formulating new objectives for unsupervised representation learning. Specifically, given an observed random variable $\mathbf{x}$ and a latent discrete variable $z$, we can express $p(\mathbf{x}|z)$, $p(z|\mathbf{x})$ and $p(z)$ in closed form in terms of a sufficiently expressive function (Eg. neural network) using our parameterization without restricting the class of these distributions. To demonstrate its usefulness, we develop two independent use cases for this parameterization: 

1. Mutual Information Maximization (MIM): MIM has become a popular means for self-supervised representation learning. Neural Bayes allows us to compute mutual information between observed random variables $\mathbf{x}$ and latent discrete random variables $z$ in closed form. We use this for learning image representations and show its usefulness on downstream classification tasks.

2. Disjoint Manifold Labeling: Neural Bayes allows us to formulate an objective which can optimally label samples from disjoint manifolds present in the support of a continuous distribution. This can be seen as a specific form of clustering where each disjoint manifold in the support is a separate cluster. We design clustering tasks that obey this formulation and empirically show that the model optimally labels the disjoint manifolds. {Our code is available at \url{https://github.com/salesforce/NeuralBayes}}
\end{abstract}

\section{Introduction}
Humans have the ability to automatically categorize objects and entities through sub-consciously defined notions of similarity, and often in the absence of any supervised signal. For instance, studies have shown that young infants are capable of automatically forming categories based on gender \cite{younger1999parsing,johnston2001developmental}, types of animals \cite{gopnik1987development,bornstein2010development}, shapes \cite{smith2006attentional}, etc. It is generally hypothesized that such discrete categorizations result in efficient encoding of sensory input that reduces the amount of information processing required by the brain \cite{rakison2010infant}. Therefore, unsupervised categorization can be seen as a means of learning useful encoding for real world data. This skill is extremely valuable since the majority of data available in the real world is unlabeled.

In this spirit, we introduce a generic parameterization that allows learning representations from unlabeled data by categorizing them.
Specifically, our parameterization implicitly maps samples from an observed random variable $\mathbf{x}$ to a latent discrete space $z$ where the distribution $p(\mathbf{x})$ gets segmented into a finite number of arbitrary conditional distributions. Imposing different conditions on the latent space $z$ through different objective functions will result in learning qualitatively different representations.

We note that our parameterization may be used to compute statistical quantities involving observed variables and latent discrete variables that are in general difficult to compute, thus providing a flexible framework for unsupervised representation learning. To illustrate this aspect, we develop two independent use cases for this parameterization-- mutual information maximization \citep{linsker1988self} and disjoint manifold labeling, as described in the abstract. For the MIM task, we experiment with benchmark image datasets and show that the unsupervised representation learned by the network achieves good performance on downstream classification tasks. For the manifold labeling task, we show experiments on 2D datasets and their high-dimensional counter-parts designed as per the problem formulation, and show that the proposed objective can optimally label disjoint manifolds. For both objectives we design regularizations necessary to achieve the desired behavior in practice.

The paper is organized as follows. We introduce the parameterization in section \ref{sec_reparam_trick}. We then develop the two applications of the parameterization, viz, mutual information maximization and disjoint manifold labeling, in section \ref{sec_mi} and section \ref{sec_manifold_iden} respectively. Finally we show experiments in section \ref{sec_exp} followed by related work and conclusion. All the proofs can be found in the appendix.

\section{Neural Bayes}
\label{sec_reparam_trick}

Consider a data distribution $p(\mathbf{x})$ from which we have access to i.i.d. samples $\mathbf{x} \in \mathbb{R}^n$. We suppose that this marginal distribution is a union of $K$ conditionals where the $k^{th}$ density is denoted by $p(\mathbf{x}|z=k)$ $\in \mathbb{R^+}$ and the corresponding probability mass denoted by $p(z=k)$ $\in \mathbb{R^+}$. Here $z$ is a discrete random variable with $K$ states. 
We now introduce the parameterization that allows us to implicitly factorize any marginal distribution into conditionals as described above. Aside from the technical details, the key idea behind this parameterization is the Bayes' rule.

\begin{lemma}
\label{lemma_reparam_trick}
Let $p(\mathbf{x}|z=k)$ and $p(z)$ be any conditional and marginal distribution defined for continuous random variable $\mathbf{x}$ and discrete random variable $z$. If $\mathbb{E}_{\mathbf{x}\sim p(\mathbf{x})}[L_k(\mathbf{x})] \neq 0$ $\forall k \in [K]$, then there exists a non-parametric function $L(\mathbf{x}): \mathbb{R}^n \rightarrow \mathbb{R}^{+^{K}}$ for any given input $\mathbf{x} \in \mathbb{R}^n$ with the property $\sum_{k=1}^K L_k(\mathbf{x})=1$ $\forall \mathbf{x}$ such that,
\begin{align}
    p(\mathbf{x}|z=k) &= \frac{L_k(\mathbf{x}) \cdot p(\mathbf{x})}{\mathbb{E}_{\mathbf{x}\sim p(\mathbf{x})}[L_k(\mathbf{x})]}, \mspace{20mu} p(z=k) = \mathbb{E}_{\mathbf{x}}[L_k(\mathbf{x})] \nonumber\\
    p(z=k|\mathbf{x})&= L_k(\mathbf{x})
\end{align}
and this parameterization is consistent.
\end{lemma}

Thus the function $L$ can be seen as a form of soft categorization of input samples. In practice, we use a neural network with sufficient capacity and softmax output to realize this function $L$. We name our parameterization method \textit{Neural Bayes} and replace $L$ with $L_\theta$ to denote the parameters of the network. By imposing different conditions on the structure of $z$ by formulating meaningful objectives, we will get qualitatively different kinds of factorization of the marginal $p(\mathbf{x})$, and therefore the function $L_\theta$ will encode the posterior for that factorization. In summary, if one formulates any objective that involves the terms $p(\mathbf{x}|z)$, $p(z)$ or $p(z|\mathbf{x})$, where $\mathbf{x}$ is an observed random variable and $z$ is a discrete latent random variable, then they can be substituted with $\frac{L_k(\mathbf{x}) \cdot p(\mathbf{x})}{\mathbb{E}_{\mathbf{x}}[L_k(\mathbf{x})]}$, $\mathbb{E}_{\mathbf{x}}[L_k(\mathbf{x})]$ and $L_k(\mathbf{x})$ respectively. 

On an important note, Neural Bayes parameterization requires using the term $\mathbb{E}_{\mathbf{x}}[L_k(\mathbf{x})]$, through which computing gradient is infeasible in general. A general discussion around this can be found in appendix \ref{sec_grad_limitation}. Nonetheless, we show that mini-batch gradients can have good fidelity for one of the objectives we propose using our parameterization. In the next two sections, we explore two different ways of factorizing $p(\mathbf{x})$ resulting in qualitatively different goals of unsupervised representation learning.

\section{Mutual Information Maximization (MIM)}
\label{sec_mi}
\subsection{Theory}
Suppose we want to find a discrete latent representation $z$ (with $K$ states) for the distribution $p(\mathbf{x})$ such that the mutual information $MI(\mathbf{x},z)$ is maximized \citep{linsker1988self}. Such an encoding $z$ demands that it must be very efficient since it has to capture maximum possible information about the continuous distribution $p(\mathbf{x})$ in just $K$ discrete states. Assuming we can learn such an encoding, we are interested in computing $p(z|\mathbf{x})$ since it tells us the likelihood of $\mathbf{x}$ belonging to each discrete state of $z$, thereby performing soft categorization which may be useful for downstream tasks. 
In the proposition below, we show an objective for computing $p(z|\mathbf{x})$ for a discrete latent representation $z$ that maximizes $MI(\mathbf{x},z)$.

\begin{proposition} (Neural Bayes-MIM-v1)
\label{theorem_mi}
Let $L(\mathbf{x}): \mathbb{R}^n \rightarrow \mathbb{R}^{+^{K}}$ be a non-parametric function for any given input $\mathbf{x} \in \mathbb{R}^n$ with the property $\sum_{i=1}^K L_k(\mathbf{x})=1$ $\forall \mathbf{x}$. Consider the following objective,
\begin{align}
\label{eq_mi_obj}
    L^* = \arg\max_{L} \mathbb{E}_{\mathbf{x}} \left[ \sum_{k=1}^K L_k(\mathbf{x}) \log \frac{L_k(\mathbf{x})}{\mathbb{E}_{\mathbf{x}}[L_k(\mathbf{x})]} \right]
\end{align}
Then $L^*_k(\mathbf{x}) = p(z^*=k|\mathbf{x})$, where $ z^* \in \arg\max_{z} MI(\mathbf{x},z)$.
\end{proposition}
The proof essentially involves expressing MI in terms of $p(z|\mathbf{x})$, and $p(z)$, which can be substituted using Neural Bayes parameterization.
However, the objective proposed in the above theorem poses a challenge-- the objective contains the term $\mathbb{E}_{\mathbf{x}}[L_k(\mathbf{x})]$ for which computing high fidelity gradient in a batch setting is problematic (see appendix \ref{sec_grad_limitation}). However, we can overcome this problem for the MIM objective because it turns out that gradient through certain terms are 0 as shown by the following theorem.

\begin{theorem}
\label{theorem_grad_block_main}
(Gradient Simplification) Denote,
\begin{align}
    J(\theta) = -\mathbb{E}_{\mathbf{x}} \left[ \sum_{k=1}^K L_{\theta_k}(\mathbf{x}) \log \frac{L_{\theta_k}(\mathbf{x})}{\mathbb{E}_{\mathbf{x}}[L_{\theta_k}(\mathbf{x})]} \right]
\end{align}
\begin{align}
\label{eq_j_hat_mim}
    \hat{J}(\theta) =  -\mathbb{E}_{\mathbf{x}} \left[ \sum_{k=1}^K L_{\theta_k}(\mathbf{x}) \log \langle \frac{L_{\theta_k}(\mathbf{x})}{\mathbb{E}_{\mathbf{x}}[L_{\theta_k}(\mathbf{x})]} \rangle \right]
\end{align}
where $\langle.\rangle$ indicates that gradients are not computed through the argument. Then $\frac{\partial J(\theta)}{\partial \theta} = \frac{\partial \hat{J}(\theta)}{\partial \theta}$.
\end{theorem}
The above theorem implies that as long as we plugin a decent estimate of $\mathbb{E}_{\mathbf{x}}[L_{\theta_k}(\mathbf{x})]$ in the objective, unbiased gradients can be computed without the need to compute gradients using the entire dataset. Note that the objective can be re-written as,
\begin{align}
\label{eq_decoupled_MIMv1}
    \min_{\theta} &-\mathbb{E}_{\mathbf{x}} \left[ \sum_{k=1}^K L_{\theta_k}(\mathbf{x}) \log \langle {L_{\theta_k}(\mathbf{x})} \rangle \right] \nonumber\\
    & + \sum_{k=1}^K \mathbb{E}_{\mathbf{x}}[L_k(\mathbf{x})] \log \langle \mathbb{E}_{\mathbf{x}}[L_k(\mathbf{x})] \rangle
\end{align}

The second term is the negative entropy of the discrete latent representation $p(z=k) := \mathbb{E}_{\mathbf{x}}[L_k(\mathbf{x})]$ which acts as a uniform prior. In other words, this term encourages learning a latent code $z$ such that all states of $z$ activate uniformly over the marginal input distribution $\mathbf{x}$. This is an attribute of distributed representation which is a fundamental goal in deep learning. We can therefore further encourage this behavior by treating the coefficient of this term as a hyper-parameter. In our experiments we confirm both the distributed representation behavior of this term as well as the benefit of using a hyper-parameter as our coefficient.

\subsection{Implementation Details}

\textbf{Alternative Formulation of Uniform Prior}: In practice we found that an alternative formulation of the second term in Eq \ref{eq_decoupled_MIMv1} results in better performance and more interpretable filters. Specifically, we replace it with the following cross-entropy formulation,
\begin{align}
\label{eq_entropy_v2}
    \mathcal{R}_p(\theta)&:= -\sum_{k=1}^K \frac{1}{K} \log  ( \mathbb{E}_{\mathbf{x}}[L_k(\mathbf{x})]) \nonumber\\
    &+  \frac{K-1}{K}\log (1- \mathbb{E}_{\mathbf{x}}[L_k(\mathbf{x})] )
\end{align}
While both, the second term in Eq \ref{eq_decoupled_MIMv1} as well as $\mathcal{R}_p(\theta)$ are minimized when $\mathbb{E}_{\mathbf{x}}[L_k(\mathbf{x})] = 1/K$, the latter formulation provides much stronger gradients during optimization when $\mathbb{E}_{\mathbf{x}}[L_k(\mathbf{x})]$ approaches 1 (see appendix \ref{sec_uniform_prior} for details); $\mathbb{E}_{\mathbf{x}}[L_k(\mathbf{x})]=1$ is undesirable since it discourages distributed representation. Finally, unbiased gradients can be computed through Eq \ref{eq_entropy_v2} as long as a good estimate of $\mathbb{E}_{\mathbf{x}}[L_k(\mathbf{x})]$ is plugged in. Also note that the condition $\mathbb{E}_{\mathbf{x}}[L_k(\mathbf{x})] \notin \{0,1\}$ in lemma \ref{lemma_reparam_trick} is met by the Neural Bayes-MIM objective implicitly during optimization as discussed in the above paragraph in regards to distributed representation.

\textbf{Implementation}: The final Neural Bayes-MIM-v2 objective is,
\begin{align}
\label{eq_MIMv2}
    \min_{\theta} &-\mathbb{E}_{\mathbf{x}} \left[ \sum_{k=1}^K L_{\theta_k}(\mathbf{x}) \log \langle {L_{\theta_k}(\mathbf{x})} +\epsilon \rangle \right] \nonumber\\
    & + (1+\alpha) \cdot \mathcal{R}_p(\theta) + \beta \cdot \mathcal{R}_c
\end{align}
where $\alpha$ and $\beta$ are hyper-parameters, $\mathcal{R}_c$ is a smoothness regularization introduced in section \ref{sec_dml_impl}, $\epsilon=10^{-7}$ is a small scalar used to prevent numerical instability. Qualitatively, we find that the regularization $\mathcal{R}_c$ prevents filters from memorizing the input samples. Finally, we apply the first two terms in Eq \ref{eq_MIMv2} to all hidden layers of a deep network at different scales (computed by spatially average pooling and applying Softmax). These two regularizations gave a significant performance boost. Thorough implementation details are provided in appendix \ref{sec_lagnet_mim_applying}. For brevity, we refer to our final objective as Neural Bayes-MIM in the rest of the paper.

On the other hand, to compute a good estimate of gradients, we use the following trick. During optimization, we compute gradients using a sufficiently large mini-batch of size MBS (Eg. $500$) that fits in memory (so that the estimate of $\mathbb{E}_{\mathbf{x}}[L_k(\mathbf{x})]$ is reasonable), and accumulate these gradients until BS samples are seen (Eg. 2000), and averaged before updating the parameters to further reduce estimation error.

\section{Disjoint Manifold Labeling (DML)}
\label{sec_manifold_iden}

\subsection{Theory}
A distribution is defined over a support. In many cases, the support may be a set of disjoint manifolds. In this task, our goal is to label samples from each disjoint manifold with a distinct value. This formulation can be seen as a generalization of subspace clustering \citep{ma2008estimation} where affine manifolds are considered. To make the problem concrete, we first formalize the definition of a disjoint manifold. 
\begin{definition}
\label{def_conn_set}
(\textit{Connected Set}) We say that a set $S \subset \mathbb{R}^n$ is a connected set (disjoint manifold) if for any $\mathbf{x}, \mathbf{y} \in S$, there exists a continuous path between $\mathbf{x}$ and $\mathbf{y}$ such that all the points on the path also belong to $S$.
\end{definition}
To identify such disjoint manifolds in a distribution, we exploit the observation that only partitions that separate one disjoint manifold from others have high divergence between the respective conditional distributions while partitions that cut through a disjoint manifold result in conditional distributions with low divergence between them. Therefore, the objective we propose for this task is to partition the unlabeled data distribution $p(\mathbf{x})$ into conditional distributions $q_i(\mathbf{x})$'s such that a divergence between them is maximized. By doing so we recover the conditional distributions defined over the disjoint manifolds (we prove its optimality in theorem \ref{theorem_dml_optimal}). We begin with two disjoint manifolds and extend this idea to multiple disjoint manifolds in appendix \ref{sec_lagnet_dml_extension}.

Let $J$ be a symmetric divergence (Eg. Jensen-Shannon divergence, Wasserstein divergence, etc), and $q_0$ and $q_1$ be the disjoint conditional distributions that we want to learn. Then the aforementioned objective can be written formally as follows:

\begin{align}
\label{eq_primary_objective}
    &\max_{q_0, q_1 \atop \pi \in (0,1)} J(q_0(\mathbf{x}) || q_1(\mathbf{x}))\\
    &\text{s.t.} \mspace{30mu}  \int_{\mathbf{x}} q_0(\mathbf{x})=1, \mspace{30mu} \int_{\mathbf{x}} q_1(\mathbf{x})=1 \nonumber\\
    & \mspace{50mu} q_1(\mathbf{x})\cdot \pi + q_0(\mathbf{x})\cdot (1-\pi) = p(\mathbf{x}).\nonumber
\end{align}

Since our goal is to simply assign labels to data samples $\mathbf{x}$ corresponding to which manifold they belong instead of learning conditional distributions as achieved by Eq. (\ref{eq_primary_objective}), we would like to learn a function $L(\mathbf{x})$ which maps samples from disjoint manifolds to distinct labels. To do so, below we derive an objective equivalent to Eq. (\ref{eq_primary_objective}) that learns such a function $L(\mathbf{x})$.
\begin{proposition}
\label{main_theorem} (Neural Bayes-DML)
Let $L(\mathbf{x}): \mathbb{R}^n \rightarrow [0,1]$ be a non-parametric function for any given input $\mathbf{x} \in \mathbb{R}^n$, and let $J$ be the Jensen-Shannon divergence. Define scalars $f_1(\mathbf{x}) := 
\frac{L(\mathbf{x})}{\mathbb{E}_{\mathbf{x}}[L(\mathbf{x})]}$ and $f_0(\mathbf{x}) := 
\frac{1-L(\mathbf{x})}{1-\mathbb{E}_{\mathbf{x}}[L(\mathbf{x})]}$. Then the objective in Eq. (\ref{eq_primary_objective}) is equivalent to,
\begin{align}
\label{eq_binary_obj_theorem}
    \max_{L} &\frac{1}{2} \cdot \mathbb{E}_{\mathbf{x}} \left[ f_1(\mathbf{x}) \cdot \log \left( \frac{f_1(\mathbf{x})}{f_1(\mathbf{x}) + f_0(\mathbf{x})} \right) \right] + \\
    &\frac{1}{2} \cdot \mathbb{E}_{\mathbf{x}} \left[f_0(\mathbf{x}) \cdot \log \left( \frac{f_0(\mathbf{x})}{f_1(\mathbf{x}) + f_0(\mathbf{x})} \right) \right] + \log 2  \nonumber\\
    & s.t. \mspace{20mu}  \mathbb{E}_{\mathbf{x}}[L(\mathbf{x})] \notin \{0,1\}.
\end{align}
\end{proposition}

\textbf{Optimality}: We now prove the optimality of the proposed objective towards discovering disjoint manifolds present in the support of a probability density function $p(\mathbf{x})$. 
\begin{theorem}
\label{theorem_dml_optimal} (optimality)
Let $p(\mathbf{x})$ be a probability density function over $\mathbb{R}^n$ whose support is the union of two non-empty connected sets (definition \ref{def_conn_set}) $S_1$ and $S_2$ that are disjoint, i.e. $S_1 \cap S_2 \ = \varnothing$. Let $L(\mathbf{x}) \in [0,1]$ belong to the class of continuous functions which is learned by solving the objective in Eq. (\ref{eq_binary_obj_theorem}). Then the objective in Eq. (\ref{eq_binary_obj_theorem}) is maximized if and only if one of the following is true:
\begin{align*}
L(\mathbf{x}) =
    \begin{cases}
    0 & \forall \mathbf{x} \in S_1 \\
    1 & \forall \mathbf{x} \in S_2
    \end{cases} \mspace{30mu} \text{or} \mspace{30mu}  L(\mathbf{x}) =
    \begin{cases}
    1 & \forall \mathbf{x} \in S_1 \\
    0 & \forall \mathbf{x} \in S_2.
    \end{cases}
\end{align*}
\end{theorem}
The above theorem proves that optimizing the derived objective over the space of functions $L$ implicitly partitions the data distribution into maximally separated conditionals by assigning a distinct label to points in each manifold. Most importantly, the theorem shows that the continuity condition on the function $L(\mathbf{x})$ plays an important role. Without this condition, the network cannot identify disjoint manifolds.


\subsection{Implementation Details}
\label{sec_dml_impl}

\textbf{Prior Collapse}: The constraint in proposition \ref{main_theorem} is a boundary condition required for technical reasons in lemma \ref{lemma_reparam_trick}. In practice we do not worry about them because optimization itself avoids situations where $\mathbb{E}_{\mathbf{x}}[L(\mathbf{x})] \in \{0,1\}$. To see the reason behind this, note that except when initialized in a way such that $\mathbb{E}_{\mathbf{x}}[L(\mathbf{x})] \in \{0,1\}$, the log terms are negative by definition. Since the denominators of $f_0$ and $f_1$ are $\mathbb{E}_{\mathbf{x}}[L(\mathbf{x})]$ and $1-\mathbb{E}_{\mathbf{x}}[L(\mathbf{x})]$ respectively, the objective is maximized when $\mathbb{E}_{\mathbf{x}}[L(\mathbf{x})]$ moves away from 0 and 1. Thus, for any reasonable initialization, optimization itself pushes $\mathbb{E}_{\mathbf{x}}[L(\mathbf{x})]$ away from 0 and 1. 



\textbf{Smoothness} of $L_\theta(.)$: As shown in theorem \ref{theorem_dml_optimal}, the proposed objectives can optimally recover disjoint manifolds only when the function $L_\theta(.)$ is continuous. In practice we found enforcing the function to be smooth (thus also continuous) helps significantly. Therefore, after experimenting with a handful of heuristics for regularizing $L_\theta$, we found the following finite difference Jacobian regularization to be effective ($L(.)$ can be scalar or vector),
\begin{align}
    \mathcal{R}_c =  \frac{1}{B}\sum_{i=1}^B \frac{ \lVert L_\theta(\mathbf{x}_i) - L_\theta(\mathbf{x}_i + \zeta \cdot \hat{\delta}_i) \rVert^2 }{\zeta^2} 
\end{align}
where $\hat{\delta}_i := \frac{\delta_i}{\lVert \delta_i \rVert_2}$ is a normalized noise vector computed independently for each sample $\mathbf{x}_i$ in a batch of size $B$ as,
\begin{align}
    \delta_i := \mathbf{X} \mathbf{v}_i.
\end{align}
Here $\mathbf{X} \in \mathbb{R}^{n \times B}$ is the matrix containing the batch of samples, and each dimension of $\mathbf{v}_i \in \mathbb{R}^B$ is sampled i.i.d. from a standard Gaussian. This computation ensures that the perturbation lies in the span of data, which we found to be important. Finally $\zeta$ is the scale of normalized noise added to all samples in a batch. In our experiments, since we always normalize the datasets to have zero mean and unit variance across all dimensions, we sample $\zeta \sim \mathcal{N}(0,0.1^2)$.

\textbf{Implementation}: We implement the binary-partition Neural Bayes-DML using the Monte-Carlo sampling approximation of the following objective,
\begin{align}
    \min_{\theta} &\frac{1}{2} \cdot \mathbb{E}_{\mathbf{x}} \left[ f_1(\mathbf{x}) \cdot \log \left(1+ \frac{f_0(\mathbf{x})}{f_1(\mathbf{x})} \right) \right] + \nonumber\\
    &\frac{1}{2} \cdot \mathbb{E}_{\mathbf{x}} \left[f_0(\mathbf{x}) \cdot \log \left(1+ \frac{f_1(\mathbf{x})}{f_0(\mathbf{x})} \right) \right] + \beta \cdot \mathcal{R}_c
\end{align}
where $f_1(\mathbf{x}) := 
\frac{L_\theta(\mathbf{x})}{\mathbb{E}_{\mathbf{x}}[L_\theta(\mathbf{x})]} + \epsilon$ and $f_0(\mathbf{x}) := 
\frac{1-L_\theta(\mathbf{x})}{1-\mathbb{E}_{\mathbf{x}}[L_\theta(\mathbf{x})]}+ \epsilon$. Here $\epsilon=10^{-7}$ is a small scalar used to prevent numerical instability, and $\beta$ is a hyper-parameter to control the continuity of $L$. The multi-partition case can be implemented in a similar way. Due to the need for computing $\mathbb{E}_{\mathbf{x}}[L_\theta(\mathbf{x})]$ in the objective, optimizing it using gradient descent methods with small batch-sizes is not possible. Therefore we experiment with this method on datasets where gradients can be computed for a very large batch-size needed to approximate the gradient through $\mathbb{E}_{\mathbf{x}}[L_\theta(\mathbf{x})]$ sufficiently well.

\section{Experiments}
\label{sec_exp}

\subsection{Mutual Information Maximization}
Instead of aiming for state-of-the-art results, our goal in this section is to conduct a preliminary (but thorough) set of experiments using Neural Bayes-MIM to understand the behavior of the algorithm, the hyper-parameters involved and do a fair comparison with popular existing methods for self-supervised learning. Therefore, we use the following simple CNN encoder architecture\footnote{ We use the following shorthand for a) conv layer: C(number of filters, filter size, stride size, padding); b) pooling: P(kernel size, stride, padding, pool mode)} $Enc$ in our experiments:  $C(200,3,1,0)-P(2,2,0,\text{max})-C(500,3,1,0)-C(700,3,1,0)-P(2,2,0,\text{max})-C(1000,3,1,0)$. For an input image $\mathbf{x}$ of size $32 \times 32 \times 3$, the output of this encoder $Enc(\mathbf{x})$ has size  $2 \times 2 \times 1000$. The encoder is initialized using orthogonal initialization \citep{saxe2013exact}, batch normalization \citep{ioffe2015batch} is used after each convolution layer and ReLU non-linearities are used. All datasets are normalized to have dimension-wise 0 mean and unit variance. Early stopping in all experiments is done using the test set (following previous work). We broadly follow the experimental setup of \citet{hjelm2018learning}. We do not use any data augmentation in our experiments. After training the encoder, we freeze its features and train a 1 hidden layer (200 units) classifier to get the final test accuracy. Extending the algorithm to more complex architectures (Eg. ResNets \cite{he2016deep}), use of multiple data augmentation techniques and other advanced regularizations (Eg. see \citet{bachman2019learning}) is left as future work.

\begin{figure}[t]
\centering
 
\subfloat[$\alpha=0$, $\beta=4$]{
	\includegraphics[trim={0cm 5.9cm 0 1.2cm},clip, width=1\linewidth]{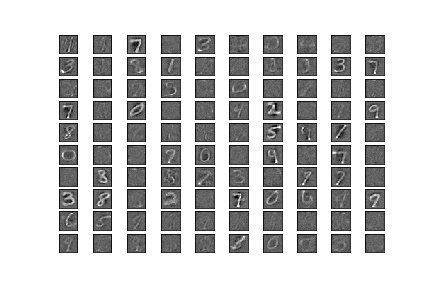} } 
 
\subfloat[$\alpha=4$, $\beta=0$]{
	\includegraphics[trim={0cm 5.9cm 0 1.2cm},clip, width=1\linewidth]{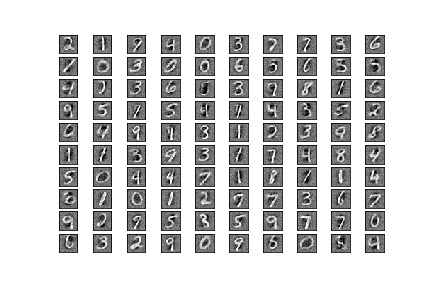} } 
 
\subfloat[$\alpha=4$, $\beta=4$]{
	\includegraphics[trim={0cm 5.9cm 0 1.2cm},clip, width=1\linewidth]{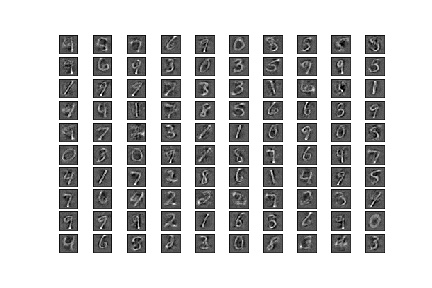} } 
 
\caption{MNIST filters learned using Neural Bayes-MIM objective Eq \ref{eq_MIMv2}. A majority of filters are dead when regularization coefficient $\alpha=0$. Filters memorize input samples when regularization coefficient $\beta=0$. Using both regularization terms results in filters that mainly capture parts of inputs which are good for distributed representations. See figure \ref{fig:mnist_filters_full} (appendix) for full images.}
\label{fig:mnist_filters}
\end{figure}

\begin{figure*}[t]
\centering
  \includegraphics[trim={0cm 1.7cm 0 5cm},clip, width=1\linewidth]{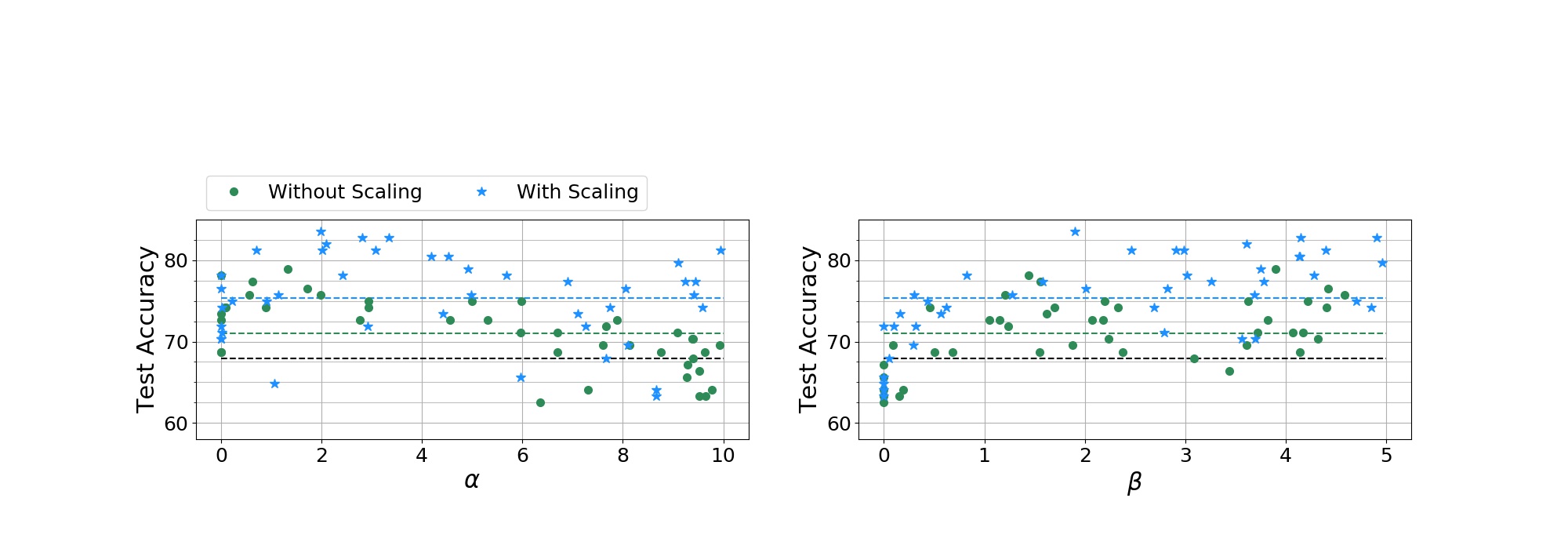}
  \caption{The effect of Neural Bayes-MIM hyper-parameters on the final test performance for CIFAR-10. A CNN encoder is trained using Neural Bayes-MIM with different configurations-- hyper-parameters $\alpha$, $\beta$ and scaling of states on which the objective is applied. A one hidden layer classifier (with 200 units) is then trained using labels on these frozen features to get the final test accuracy. The plots show that performance is significantly worse when $\alpha=\beta=0$ and no scaling is used, showing their important role as a regularizer. The best performing model reaches $83.59\%$. Black dotted line is the baseline performance ($67.97\%$) when a randomly initialized network (with identical architecture) is used as encoder. Green and blue dotted lines are the average of all the green and blue points respectively.}
  \label{fig:c10_3ablation}
\end{figure*}

\subsubsection{Ablation Studies}

\textbf{Behavior of Neural Bayes-MIM-v1 (Eq \ref{eq_decoupled_MIMv1}) vs Neural Bayes-MIM (v2, Eq \ref{eq_entropy_v2})}: The experiments and details are discussed in appendix \ref{sec_v1_vs_v2}. The main differences are: 1. majority of the filters learned by the v1 objective are dead, as opposed to the v2 objective which encourages distributed representation; 2. the performance of v2 is better than that of the v1 objective.

\textbf{Visualization of Filters}: We visualize the filters learned by the Neural Bayes-MIM objective on MNIST digits and qualitatively study the effects of the regularizations used. For this we train a deep fully connected network with 3 hidden layers each of width 500 using Adam with learning rate 0.001, batch size 500, 0 weight decay for 50 epochs (other Adam hyper-parameters are kept standard). We train three configurations: 1. $\alpha=0$, $\beta=4$; 2. $\alpha=4$, $\beta=0$; 3. $\alpha=4$, $\beta=4$. The learned filters are shown in figure \ref{fig:mnist_filters}. We find that the uniform prior regularization ($\alpha>0$) prevents dead filters while the smoothness regularization ($\beta>0$) prevents input memorization.

\textbf{Performance due to Regularizations and State Scaling}: We now evaluate the effects of the various components involved in the Neural Bayes-MIM objective-- coefficients $\alpha$ and $\beta$, and applying the objective at different scales of hidden states. We use the CIFAR-10 dataset for these experiments.

In the first experiment, for each value of the number of different scales considered, we vary $\alpha$, $\beta$ and record the final performance, thus capturing the variation in performance due to all these three components. We consider two scaling configurations: 1. no pooling is applied to the hidden layers; 2. for each hidden layer, we spatially average pool the state using a $2 \times 2$ pooing filter with a stride of 2. For the encoder used in our experiments (which has 4 internal hidden layers post ReLU), this gives us 4 and 8 states respectively (including the original un-scaled hidden layers) to apply the Neural Bayes-MIM objective. After getting all the states, we apply the Softmax activation to each state along the channel dimension so that the Neural Bayes parameterization holds. Thus for states with height and width, the objective is applied to each spatial $(x,y)$ location separately and averaged. Also, for states with height (or width) less than the pooling size, we use the height (or width) as pooling size.

We train Neural Bayes-MIM on the full training set for 100 epochs using Adam with learning rate 0.001 (other Adam hyper-parameters are standard), mini-batch size 500 and batch size 2000, 0 weight decay. In the first 32 experiments, $\alpha$ and $\beta$ are sampled uniformly from $[0,10]$ and $[0,5]$ respectively. In the next 5 experiments, $\alpha$ is set to be 0 while $\beta$ is sampled uniformly. In the next 5 experiments, $\beta$ is set to be 0 while $\alpha$ is sampled uniformly. Thus in total we run 42 experiments for each number of scaling considered.

Once we get a trained $Enc(\mathbf{x})$, we train a 1 hidden layer (with 200 units) MLP classifier on the frozen features from $Enc(\mathbf{x})$ using the labels in the training set. This training is done for 100 epochs using Adam with learning rate 0.001 (other Adam hyper-parameters are standard), batch size 128 and weight decay 0.

As a baseline for these experiments, we use a randomly initialed encoder $Enc(\mathbf{x})$. Since there are no tunable hyper-parameters in this case, we perform a grid search on the classifier hyper-parameters. Specifically, we choose weight decay from $\{ 0, 0.00001, 0.00005, 0.0001 \}$, batch size from $\{128, 256\}$, and learning rate from $\{0.0001, 0.001\}$. This yields a total of 16 configurations. The test accuracy from these runs varied between $58.59 \%$ and $67.97 \%$. We consider $67.97 \%$ as our baseline.

The performance of encoders under the aforementioned configurations is shown in figure \ref{fig:c10_3ablation}. It is clear that both the hyper-parameters $\alpha$ and especially $\beta$ play an important role in the quality of representations learned. Also, applying Neural Bayes-MIM at different scales of the network states significantly improves the average and best performance.

\textbf{Effect of Mini-batch size (MBS) and Batch size (BS)}: During implementation, we proposed to compute gradients using a reasonably large mini-batch of size MBS and accumulate gradients until BS samples are seen. This is done to overcome the gradient estimation problem due to the $\mathbb{E}_{\mathbf{x}}[L_k(\mathbf{x})]$ term in Neural Bayes-MIM. Here we evaluate the effect of these two hyper-parameters on the final test performance. We choose MBS from $\{ 50,100,250,500 \}$ and BS from $\{ 50,250,500,2000,3000\}$. For each combination of MBS and BS, we train the CNN encoder using Neural Bayes-MIM with $\alpha=2$ and $\beta=4$ (chosen by examining figure \ref{fig:c10_3ablation}); the rest of the training settings are kept identical to those used for figure \ref{fig:c10_3ablation} experiment. Table \ref{table_bs_ablation} shows the final test accuracy on CIFAR-10 for each combination of hyper-parameters MBS and BS. We make two observations: 1. using very small MBS (Eg. 50 and 100) typically results in poor (even worse than that of a random encoder ($67.97\%$)), while larger MBS significantly improves performance; 2. using a larger BS further improves performance in most cases (even when MBS is small).

\begin{figure}[t]
\centering
\includegraphics[trim={0cm 0cm 0 1.2cm},clip, width=1\linewidth]{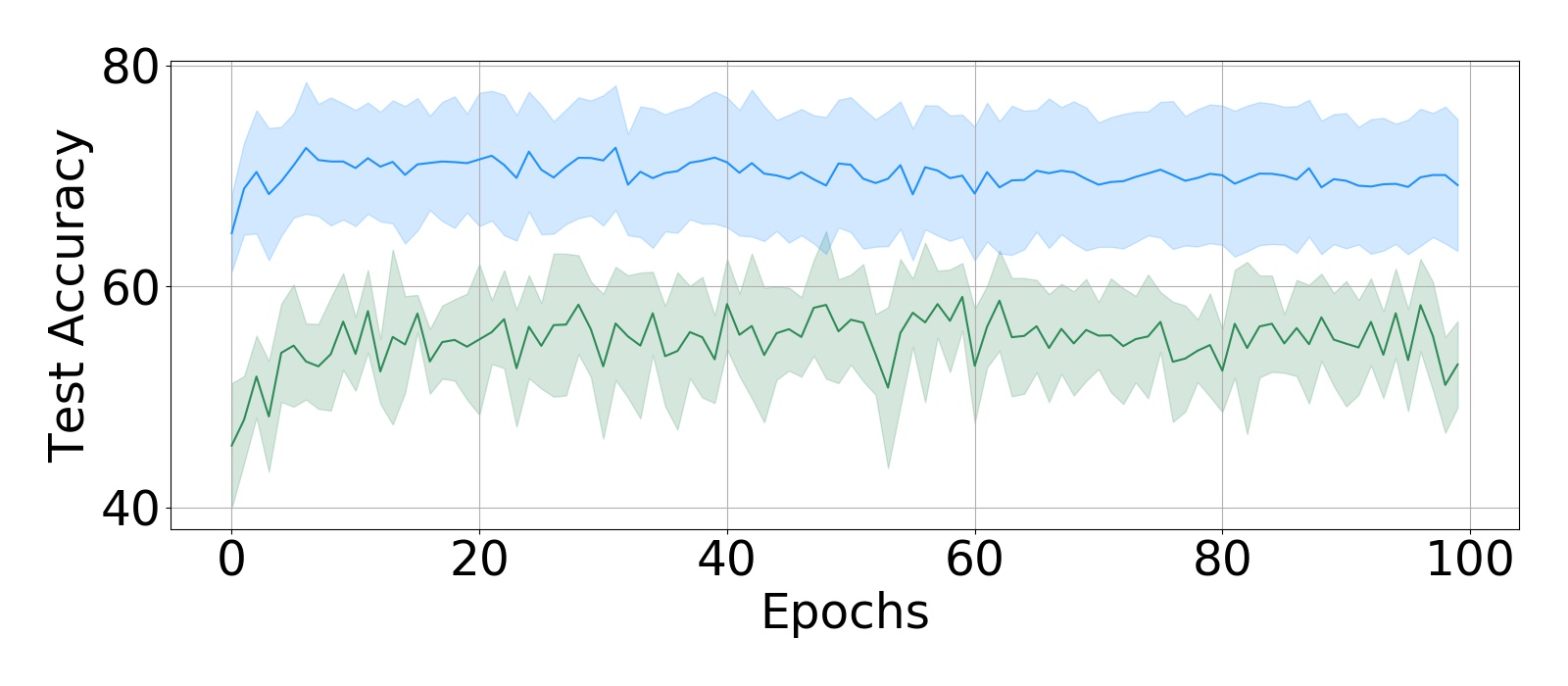} 
\caption{Mean (solid line) and standard deviation (error band) of accuracy evolution during MLP classifier training on CIFAR-10 using Neural Bayes-MIM encoder (blue) and random encoder (green). Features learned by Neural Bayes-MIM allow training to start at a much higher value ($\sim 65\%$ on average) and converge faster as opposed to a random encoder ($\sim 47\%$ on average).}
\label{fig:c10_convergence}
\end{figure}

\textbf{Accuracy vs Epochs}: Finally, we plot the evolution of accuracy over epochs for all the models learned in the experiments of figure \ref{fig:c10_3ablation}. For Neural Bayes-MIM we use the models with scaling (42 in total), and all 16 models for the random encoder. The convergence plot is shown in figure \ref{fig:c10_convergence}.

\begin{table}
{\small
\begin{center}
 \begin{tabular}{|c|| c | c | c| c| c|} 
 \hline
 MBS \textbackslash BS & 50 & 250 & 500 & 2000 & 3000 \\ [0.5ex] 
 \hline\hline
 50 & 40.62 & 42.97 & 41.41 & 75 & 78.91 \\
 \hline
 100 & N/A & 67.97 & 66.41 & 78.12 & 78.91 \\
 \hline
 250 & N/A & 76.56 & 78.91 & 82.03 & 84.38 \\
 \hline
  500 & N/A & N/A & 82.03 & 78.91 & 79.69 \\
 \hline
\end{tabular}
 \caption{\label{table_bs_ablation} Effect of mini-batch size (MBS) and batch size (BS) on final test accuracy of Neural Bayes-MIM on CIFAR-10. Gradients are computed using batches of size MBS and accumulated until BS samples are seen before parameter update. As expected, a sufficiently large MBS is needed for computing high fidelity gradients due to $\mathbb{E}_{\mathbf{x}}[L_k(\mathbf{x})]$ term. Gradient accumulation using BS further helps. All models are trained for the same number of \textit{epochs}.}
\end{center}
}
\end{table}

\subsubsection{Final Classification Performance}
We compare the final test accuracy of Neural Bayes-MIM with 3 baselines-- a random encoder (described in ablation studies), Deep Infomax \citep{hjelm2018learning}, and Rotation Prediction based representation learning \citep{gidaris2018unsupervised} on benchmark image datasets-- CIFAR10 and CIFAR-100 \citep{cifar} and STL-10 \citep{coates2011analysis}.
Random Network refers to the use of a randomly initialized network. The experimental details for them are identical to those in our ablation involving a hyper-parameter search over 16 configurations done for each dataset separately.

DIM results are reported from \citet{hjelm2018learning}. We omit STL-10 number for DIM because we resize images to a much smaller size of $32 \times 32$ in our runs instead of $64 \times 64$ as used in DIM. 

Rotation prediction refers to the algorithm in \citet{gidaris2018unsupervised} where the encoder is learned by training it to predict the rotation of unlabeled images. We use the same CNN architecture used in previous experiments, with a linear classifier added on top, and train it to predict 4 rotations angles-- 0, 90, 180, 270. We run this pre-training with 8 configurations of hyper-parameters-- batch-size $\in \{ 25, 50\}$ (each batch further includes rotated copies of each sample making the total batch-size 100, 200), weight decay $\in \{ 0, 0.00001 \}$ and learning rate $\in \{ 0.0001,0.001 \}$. For each run, we then train a 1 hidden layer (200 units) classifier on top of the frozen features with learning rate $\in \{ 0.0001,0.001 \}$. We report the best performance of all runs. Since \citet{kolesnikov2019revisiting} report that lower layers in CNN architectures trained with rotation prediction have better performance on downstream tasks, we also train classifiers on the $2^{nd}$ layer and report their performance which is significantly better.

The following describes the experiment details for Neural Bayes-MIM. We use $\alpha=2$ and $\beta=4$ (chosen roughly by examining figure \ref{fig:c10_3ablation}), and MBS=500, BS=4000 in all the experiments. Note these values are not tuned for STL-10 and CIFAR-100. For CIFAR-10 and STL-10 each, we run 4 configurations of Neural Bayes-MIM over hyper-parameters learning rate $\in \{ 0.0001,0.001 \}$ and weight decay $\in \{ 0, 0.00001 \}$. For each run, we then train a 1 hidden layer (200 units) classifier on top of the frozen features with learning rate $\in \{ 0.0001,0.001 \}$. We report the best performance of all runs. For CIFAR-100, we take the encoder that produces the best performance on CIFAR-10, and train a classifier with the 2 learning rates and report the best of the 2 runs. Similar to rotation prediction, we also train classifiers on the $2^{nd}$ layer and report their performance.

Table \ref{table_mim_clf} reports the classification performance of all the methods. We note that all experiments were done with CNN architecture without any data augmentation.  Neural Bayes-MIM outperforms baseline methods in general. However, when using $2^{nd}$ layer features, rotation prediction (RP) performs better. We hope to further improve the performance of Neural Bayes-MIM with additional regularizations similar to \citet{bachman2019learning}.

\begin{table}
{\small
\begin{center}
 \begin{tabular}{|c| c c c|} 
 \hline
 Encoder \textbackslash Dataset & CIFAR-10 & CIFAR-100 & STL-10 \\ [0.5ex] 
 \hline\hline
 Random Network & 67.97 & 42.97 & 53.91 \\ 
 \hline
 Rotation Prediction (RP) & 33.59 & 6.25 & 25.78 \\
 \hline
 DIM \citep{hjelm2018learning} & 80.95 & 49.74 & - \\ [1ex] 
 \hline
  Neural Bayes-MIM & \textbf{82.81} & \textbf{55.47} & \textbf{64.84} \\
 \hline
  \hline
 RP ($2^{nd}$ layer) & \textbf{84.38} & \textbf{64.06} & \textbf{70.31} \\
  \hline
  Neural Bayes-MIM ($2^{nd}$ layer) & \textbf{84.38} & {57.03} & 67.19 \\
 \hline
\end{tabular}
 \caption{\label{table_mim_clf}Classification performance of a one hidden layer MLP classifier trained on frozen features from the mentioned encoder models (trained using Neural Bayes-MIM) and datasets (no data augmentation used in experiments we ran). STL-10 was resized to $32 \times 32$ in our runs instead of $64 \times 64$ as in DIM due to memory restrictions. Performance reported from DIM paper are their best numbers (omitting STL-10 due to difference in image size).}
\end{center}
}
\end{table}

\begin{figure}[t]
\vspace{-0pt}
\centering
\hspace{-0.5cm}\includegraphics[trim={2cm 4cm 2cm 4cm},clip, width=0.45\linewidth]{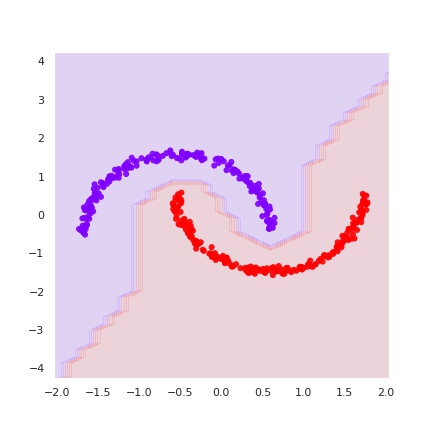}
\includegraphics[trim={2cm 3.8cm 1.8cm 3.9cm},clip, width=0.44\linewidth]{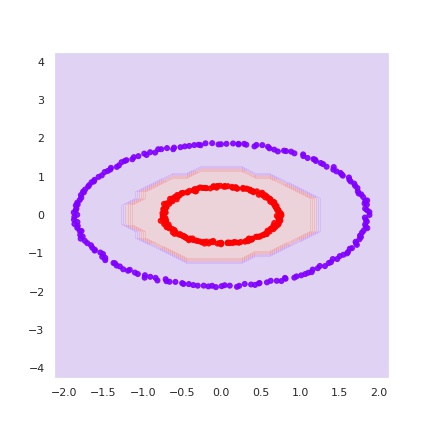}

\vspace{0.05cm}
\includegraphics[trim={2cm 3.4cm 2cm 4cm},clip, width=0.45\linewidth]{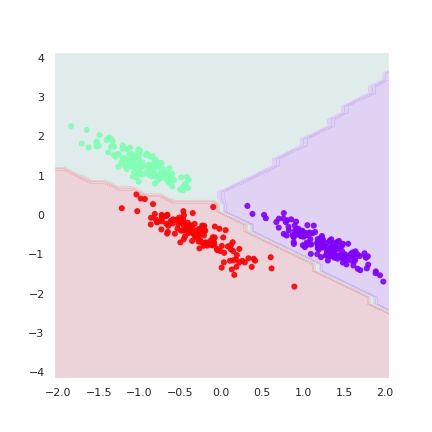}
\includegraphics[trim={2cm 3cm 0 3.9cm},clip, width=0.5\linewidth]{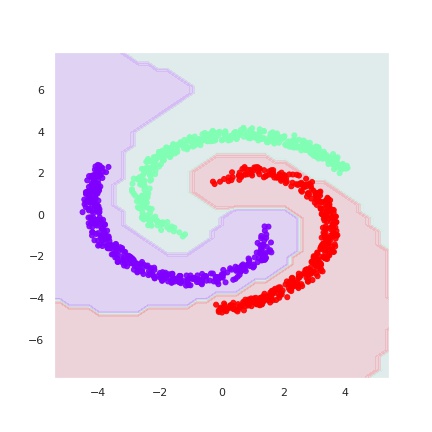}
\caption{Neural Bayes-DML network prediction on synthetic datasets. The different colors denote the label predicted by the network $L_\theta(.)$ thresholded at 0.5. The darker shades of colors denote predictions made at training data points while the lighter shades denote predictions on the rest of the space.}
\label{fig:dml_figs}
\end{figure}

\subsection{Disjoint Manifold Labeling}

Clustering in general is an ill posed problem. However, in our problem setup, the definition is precise, i.e., our goal is to optimally label all the disjoint manifolds present in the support of a distribution. Since this is a unique goal that is not generally considered in literature, as empirical verification, we show qualitative results on 2D synthetic datasets in figure \ref{fig:dml_figs}. Top 2 sub-figures have 2 clusters and the bottom 2 have 3 clusters. For all experiments we use a 4 layer MLP with 400 hidden units each, batchnorm, ReLU activation, and last layer Softmax activation. In all cases we train using Adam optimizer with a learning rate of 0.001, batch size of 400 and no weight decay, and trained until convergence. Regularization coefficient $\beta$ was chosen from $[0.5,6]$ that resulted in optimal clustering. For generality in these experiments, these 2D datasets were projected to high dimensions (512) by appending 510 dimensions of 0 entries to each sample and then randomly rotated before performing clustering. The datasets were then projected back to the original 2D space for visualizing predictions. Additional experiments can be found in appendix \ref{sec_lagnet_dml_mnist}.

\section{Related Work}
\label{sec_related_work}

Neural Bayes-MIM maximizes mutual information for learning useful representations in a self-supervised way. Introduced in \citet{linsker1988self} and \citet{bell1995information}, there are a myriad of self-supervised methods that involve MIM. As discussed in \citet{vincent2010stacked}, auto-encoder based methods achieve this goal implicitly by minimizing the reconstruction error of the input samples under isotropic Gaussian assumption. Deep infomax (DIM, \citet{hjelm2018learning}) instead uses MINE \cite{belghazi2018mine} to estimate MI and maximize it while applying it to both local and global features and imposing priors on the learned representation. \citet{hjelm2018learning} have also shown that DIM performs better than representations learned by auto-encoder based methods such as VAE \citep{kingma2013auto}, $\beta$-VAE \citep{higgins2017beta} and adversarial auto-encoder \citep{makhzani2015adversarial}, among others such as noise as targets \citep{bojanowski2017unsupervised} and BiGAN \citep{donahue2016adversarial}. Contrastive Predictive Coding \citep{oord2018representation} also maximizes MI by predicting lower layer representations from higher layers using a contrastive loss instead of reconstruction loss.

Unlike the aforementioned methods that learn continuous latent representation, Neural Bayes-MIM implicitly learns discrete latent representations. We note that the estimation of mutual information due to Neural Bayes parameterization in the Neural Bayes-MIM-v1 objective (Eq \ref{eq_decoupled_MIMv1}) turns out to be identical to the one proposed in IMSAT \citep{hu2017learning}. However, there are important differences: 1. we provide theoretical justifications for the parameterization used (lemma \ref{lemma_reparam_trick}) and show in theorem \ref{theorem_grad_block_main} why it is feasible to compute high fidelity gradients using this objective in the mini-batch setting even though it contains the term $\mathbb{E}_{\mathbf{x}}[L_k(\mathbf{x})]$. On the other hand, the justification used in IMSAT is that optimizing using mini-batches is equivalent to optimizing an upper bound of the original objective; 2. while the MI part of IMSAT was introduced in the context of clustering, we improve the MI formulation (Eq \ref{eq_MIMv2}), and introduce regularization terms and state scaling which are important for learning useful representations using the Neural Bayes-MIM objective that perform well on downstream classification tasks; 3. we perform extensive ablation studies exposing the role of the introduced regularizations; 4. the goal of our paper is broader, i.e., to introduce the Neural Bayes parameterization that can be used for formulating new objectives.
From the aspect of learning discrete latent representation, Neural Bayes-MIM has similarities with VQ-VAE \citep{oord2017neural}. However, similar to other auto-encoder based methods, VQ-VAE imposes the isotropy assumption in the reconstruction loss.

In many self-supervised methods, the idea is to learn useful representations by predicting non-trivial information about the input. Examples of such methods are Rotation Prediction \citep{gidaris2018unsupervised}, Exemplar \citep{dosovitskiy2014discriminative}, Jigsaw \citep{noroozi2016unsupervised} and Relative Patch Location \citep{doersch2015unsupervised}. \citet{kolesnikov2019revisiting} have extensively compared these methods and found that Rotation Prediction (RP) in general outperforms or performs at par with the latter methods. For the aforementioned reasons, we compared Neural Bayes-MIM with RP and DIM.

Numerous recent papers have proposed clustering algorithm for unsupervised representation learning such as Deep Clustering \citep{caron2018deep}, information based clustering \citep{ji2019invariant}, Spectral Clustering \citep{shaham2018spectralnet}, Assosiative Deep Clustering \citep{haeusser2018associative} etc. Our goal in regards to clustering in Neural Bayes-DML is in general different from such methods. Our objective is aimed at labeling disjoint manifolds in a distribution. Thus it can be seen as a generalization of the traditional subspace clustering methods \citep{ma2008estimation,liu2010robust} from affine subspaces to arbitrary manifolds.

\section{Conclusion}
We proposed a parameterization method that can be used to express an arbitrary set of distributions $p(\mathbf{x}|z)$, $p(z|\mathbf{x})$ and $p(z)$ in closed form using a neural network with sufficient capacity, which can in turn be used to formulate new objective functions. We formulated two different objectives that use this parameterization which were aimed towards different goals of self-supervised learning-- learning deep network features using the infomax principle, and identification of disjoint manifolds in the support of continuous distributions. We presented theoretical and empirical analysis of both the objectives while especially focusing on the former since it has broader applications.

\section*{Acknowledgments}
I (DA) was supported by IVADO during my time at MILA and currently supported by Salesforce. There are many people who have directly or indirectly contributed to this work and we would like to thank them. During the early phase of research on Neural Bayes-DML (in the context of which the Neural Bayes parameterization was developed), Chen Xing pointed out an intuition which led me to simplify its optimization procedure. We thank Ali Madani and Ehsan Hosseini-Asl for exploring Neural Bayes-DML for unsupervised representation learning for images. We thank Min Lin for taking interest in the connection between Neural Bayes-DML and mutual information, which led me to the idea that mutual information can be computed using the parameterization. We thank Aadyot Bhatnagar and Weiran Wang for proof-checking the paper and providing helpful feedback. We thank Devon Hjelm and Alex Fedorov for discussing their algorithm Deep Infomax in great detail. Finally, we thank Aaron Courville, Sharan Vaswani, Nikhil Naik, Isabela Albuquerque, Lav Varshney, Yu Bai, Jonathan Binas, David Krueger, Tegan Maharaj and Govardana Sachithanandam Ramachandran for helpful discussions.

\bibliography{ref}

\begin{thebibliography}{36}
\providecommand{\natexlab}[1]{#1}
\providecommand{\url}[1]{\texttt{#1}}
\expandafter\ifx\csname urlstyle\endcsname\relax
  \providecommand{\doi}[1]{doi: #1}\else
  \providecommand{\doi}{doi: \begingroup \urlstyle{rm}\Url}\fi

\bibitem[Bachman et~al.(2019)Bachman, Hjelm, and
  Buchwalter]{bachman2019learning}
Bachman, P., Hjelm, R.~D., and Buchwalter, W.
\newblock Learning representations by maximizing mutual information across
  views.
\newblock In \emph{Advances in Neural Information Processing Systems}, pp.\
  15509--15519, 2019.

\bibitem[Belghazi et~al.(2018)Belghazi, Baratin, Rajeswar, Ozair, Bengio,
  Courville, and Hjelm]{belghazi2018mine}
Belghazi, M.~I., Baratin, A., Rajeswar, S., Ozair, S., Bengio, Y., Courville,
  A., and Hjelm, R.~D.
\newblock Mine: mutual information neural estimation.
\newblock \emph{arXiv preprint arXiv:1801.04062}, 2018.

\bibitem[Bell \& Sejnowski(1995)Bell and Sejnowski]{bell1995information}
Bell, A.~J. and Sejnowski, T.~J.
\newblock An information-maximization approach to blind separation and blind
  deconvolution.
\newblock \emph{Neural computation}, 7\penalty0 (6):\penalty0 1129--1159, 1995.

\bibitem[Bojanowski \& Joulin(2017)Bojanowski and
  Joulin]{bojanowski2017unsupervised}
Bojanowski, P. and Joulin, A.
\newblock Unsupervised learning by predicting noise.
\newblock In \emph{Proceedings of the 34th International Conference on Machine
  Learning-Volume 70}, pp.\  517--526. JMLR. org, 2017.

\bibitem[Bornstein \& Arterberry(2010)Bornstein and
  Arterberry]{bornstein2010development}
Bornstein, M.~H. and Arterberry, M.~E.
\newblock The development of object categorization in young children:
  Hierarchical inclusiveness, age, perceptual attribute, and group versus
  individual analyses.
\newblock \emph{Developmental psychology}, 46\penalty0 (2):\penalty0 350, 2010.

\bibitem[Caron et~al.(2018)Caron, Bojanowski, Joulin, and Douze]{caron2018deep}
Caron, M., Bojanowski, P., Joulin, A., and Douze, M.
\newblock Deep clustering for unsupervised learning of visual features.
\newblock In \emph{Proceedings of the European Conference on Computer Vision
  (ECCV)}, pp.\  132--149, 2018.

\bibitem[Coates et~al.(2011)Coates, Ng, and Lee]{coates2011analysis}
Coates, A., Ng, A., and Lee, H.
\newblock An analysis of single-layer networks in unsupervised feature
  learning.
\newblock In \emph{Proceedings of the fourteenth international conference on
  artificial intelligence and statistics}, pp.\  215--223, 2011.

\bibitem[Doersch et~al.(2015)Doersch, Gupta, and
  Efros]{doersch2015unsupervised}
Doersch, C., Gupta, A., and Efros, A.~A.
\newblock Unsupervised visual representation learning by context prediction.
\newblock In \emph{Proceedings of the IEEE International Conference on Computer
  Vision}, pp.\  1422--1430, 2015.

\bibitem[Donahue et~al.(2016)Donahue, Kr{\"a}henb{\"u}hl, and
  Darrell]{donahue2016adversarial}
Donahue, J., Kr{\"a}henb{\"u}hl, P., and Darrell, T.
\newblock Adversarial feature learning.
\newblock \emph{arXiv preprint arXiv:1605.09782}, 2016.

\bibitem[Dosovitskiy et~al.(2014)Dosovitskiy, Springenberg, Riedmiller, and
  Brox]{dosovitskiy2014discriminative}
Dosovitskiy, A., Springenberg, J.~T., Riedmiller, M., and Brox, T.
\newblock Discriminative unsupervised feature learning with convolutional
  neural networks.
\newblock In \emph{Advances in neural information processing systems}, pp.\
  766--774, 2014.

\bibitem[Gidaris et~al.(2018)Gidaris, Singh, and
  Komodakis]{gidaris2018unsupervised}
Gidaris, S., Singh, P., and Komodakis, N.
\newblock Unsupervised representation learning by predicting image rotations.
\newblock \emph{arXiv preprint arXiv:1803.07728}, 2018.

\bibitem[Gopnik \& Meltzoff(1987)Gopnik and Meltzoff]{gopnik1987development}
Gopnik, A. and Meltzoff, A.
\newblock The development of categorization in the second year and its relation
  to other cognitive and linguistic developments.
\newblock \emph{Child development}, pp.\  1523--1531, 1987.

\bibitem[Haeusser et~al.(2018)Haeusser, Plapp, Golkov, Aljalbout, and
  Cremers]{haeusser2018associative}
Haeusser, P., Plapp, J., Golkov, V., Aljalbout, E., and Cremers, D.
\newblock Associative deep clustering: Training a classification network with
  no labels.
\newblock In \emph{German Conference on Pattern Recognition}, pp.\  18--32.
  Springer, 2018.

\bibitem[He et~al.(2016)He, Zhang, Ren, and Sun]{he2016deep}
He, K., Zhang, X., Ren, S., and Sun, J.
\newblock Deep residual learning for image recognition.
\newblock In \emph{Proceedings of the IEEE conference on computer vision and
  pattern recognition}, pp.\  770--778, 2016.

\bibitem[Higgins et~al.(2017)Higgins, Matthey, Pal, Burgess, Glorot, Botvinick,
  Mohamed, and Lerchner]{higgins2017beta}
Higgins, I., Matthey, L., Pal, A., Burgess, C., Glorot, X., Botvinick, M.,
  Mohamed, S., and Lerchner, A.
\newblock beta-vae: Learning basic visual concepts with a constrained
  variational framework.
\newblock \emph{International Conference on Learning Representations (ICLR)},
  2017.

\bibitem[Hjelm et~al.(2019)Hjelm, Fedorov, Lavoie-Marchildon, Grewal, Bachman,
  Trischler, and Bengio]{hjelm2018learning}
Hjelm, R.~D., Fedorov, A., Lavoie-Marchildon, S., Grewal, K., Bachman, P.,
  Trischler, A., and Bengio, Y.
\newblock Learning deep representations by mutual information estimation and
  maximization.
\newblock \emph{International Conference on Learning Representations (ICLR)},
  2019.

\bibitem[Hu et~al.(2017)Hu, Miyato, Tokui, Matsumoto, and
  Sugiyama]{hu2017learning}
Hu, W., Miyato, T., Tokui, S., Matsumoto, E., and Sugiyama, M.
\newblock Learning discrete representations via information maximizing
  self-augmented training.
\newblock In \emph{Proceedings of the 34th International Conference on Machine
  Learning-Volume 70}, pp.\  1558--1567. JMLR. org, 2017.

\bibitem[Ioffe \& Szegedy(2015)Ioffe and Szegedy]{ioffe2015batch}
Ioffe, S. and Szegedy, C.
\newblock Batch normalization: Accelerating deep network training by reducing
  internal covariate shift.
\newblock \emph{arXiv preprint arXiv:1502.03167}, 2015.

\bibitem[Ji et~al.(2019)Ji, Henriques, and Vedaldi]{ji2019invariant}
Ji, X., Henriques, J.~F., and Vedaldi, A.
\newblock Invariant information clustering for unsupervised image
  classification and segmentation.
\newblock In \emph{Proceedings of the IEEE International Conference on Computer
  Vision}, pp.\  9865--9874, 2019.

\bibitem[Johnston et~al.(2001)Johnston, Bittinger, Smith, and
  Madole]{johnston2001developmental}
Johnston, K.~E., Bittinger, K., Smith, A., and Madole, K.~L.
\newblock Developmental changes in infants' and toddlers' attention to gender
  categories.
\newblock \emph{Merrill-Palmer Quarterly (1982-)}, pp.\  563--584, 2001.

\bibitem[Kingma \& Welling(2013)Kingma and Welling]{kingma2013auto}
Kingma, D.~P. and Welling, M.
\newblock Auto-encoding variational bayes.
\newblock \emph{arXiv preprint arXiv:1312.6114}, 2013.

\bibitem[Kolesnikov et~al.(2019)Kolesnikov, Zhai, and
  Beyer]{kolesnikov2019revisiting}
Kolesnikov, A., Zhai, X., and Beyer, L.
\newblock Revisiting self-supervised visual representation learning.
\newblock In \emph{Proceedings of the IEEE conference on Computer Vision and
  Pattern Recognition}, pp.\  1920--1929, 2019.

\bibitem[Krizhevsky(2009)]{cifar}
Krizhevsky, A.
\newblock Learning multiple layers of features from tiny images.
\newblock Technical report, 2009.

\bibitem[Linsker(1988)]{linsker1988self}
Linsker, R.
\newblock Self-organization in a perceptual network.
\newblock \emph{Computer}, 21\penalty0 (3):\penalty0 105--117, 1988.

\bibitem[Liu et~al.(2010)Liu, Lin, and Yu]{liu2010robust}
Liu, G., Lin, Z., and Yu, Y.
\newblock Robust subspace segmentation by low-rank representation.
\newblock In \emph{Proceedings of the 27th international conference on machine
  learning (ICML-10)}, pp.\  663--670, 2010.

\bibitem[Ma et~al.(2008)Ma, Yang, Derksen, and Fossum]{ma2008estimation}
Ma, Y., Yang, A.~Y., Derksen, H., and Fossum, R.
\newblock Estimation of subspace arrangements with applications in modeling and
  segmenting mixed data.
\newblock \emph{SIAM review}, 50\penalty0 (3):\penalty0 413--458, 2008.

\bibitem[Makhzani et~al.(2015)Makhzani, Shlens, Jaitly, Goodfellow, and
  Frey]{makhzani2015adversarial}
Makhzani, A., Shlens, J., Jaitly, N., Goodfellow, I., and Frey, B.
\newblock Adversarial autoencoders.
\newblock \emph{arXiv preprint arXiv:1511.05644}, 2015.

\bibitem[Noroozi \& Favaro(2016)Noroozi and Favaro]{noroozi2016unsupervised}
Noroozi, M. and Favaro, P.
\newblock Unsupervised learning of visual representations by solving jigsaw
  puzzles.
\newblock In \emph{European Conference on Computer Vision}, pp.\  69--84.
  Springer, 2016.

\bibitem[Oord et~al.(2017)Oord, Vinyals, and Kavukcuoglu]{oord2017neural}
Oord, A. v.~d., Vinyals, O., and Kavukcuoglu, K.
\newblock Neural discrete representation learning.
\newblock \emph{arXiv preprint arXiv:1711.00937}, 2017.

\bibitem[Oord et~al.(2018)Oord, Li, and Vinyals]{oord2018representation}
Oord, A. v.~d., Li, Y., and Vinyals, O.
\newblock Representation learning with contrastive predictive coding.
\newblock \emph{arXiv preprint arXiv:1807.03748}, 2018.

\bibitem[Rakison \& Yermolayeva(2010)Rakison and
  Yermolayeva]{rakison2010infant}
Rakison, D.~H. and Yermolayeva, Y.
\newblock Infant categorization.
\newblock \emph{Wiley Interdisciplinary Reviews: Cognitive Science}, 1\penalty0
  (6):\penalty0 894--905, 2010.

\bibitem[Saxe et~al.(2013)Saxe, McClelland, and Ganguli]{saxe2013exact}
Saxe, A.~M., McClelland, J.~L., and Ganguli, S.
\newblock Exact solutions to the nonlinear dynamics of learning in deep linear
  neural networks.
\newblock \emph{arXiv preprint arXiv:1312.6120}, 2013.

\bibitem[Shaham et~al.(2018)Shaham, Stanton, Li, Nadler, Basri, and
  Kluger]{shaham2018spectralnet}
Shaham, U., Stanton, K., Li, H., Nadler, B., Basri, R., and Kluger, Y.
\newblock Spectralnet: Spectral clustering using deep neural networks.
\newblock \emph{arXiv preprint arXiv:1801.01587}, 2018.

\bibitem[Smith \& Samuelson(2006)Smith and Samuelson]{smith2006attentional}
Smith, L.~B. and Samuelson, L.
\newblock An attentional learning account of the shape bias: Reply to cimpian
  and markman (2005) and booth, waxman, and huang (2005).
\newblock \emph{Developmental Psychology}, 42\penalty0 (6):\penalty0
  1339--1343, 2006.

\bibitem[Vincent et~al.(2010)Vincent, Larochelle, Lajoie, Bengio, and
  Manzagol]{vincent2010stacked}
Vincent, P., Larochelle, H., Lajoie, I., Bengio, Y., and Manzagol, P.-A.
\newblock Stacked denoising autoencoders: Learning useful representations in a
  deep network with a local denoising criterion.
\newblock \emph{Journal of machine learning research}, 11\penalty0
  (Dec):\penalty0 3371--3408, 2010.

\bibitem[Younger \& Fearing(1999)Younger and Fearing]{younger1999parsing}
Younger, B.~A. and Fearing, D.~D.
\newblock Parsing items into separate categories: Developmental change in
  infant categorization.
\newblock \emph{Child Development}, 70\penalty0 (2):\penalty0 291--303, 1999.

\end{thebibliography}
\bibliographystyle{icml2020}

\newpage
\setcounter{page}{1}
\appendix
\onecolumn
\section*{Appendix}

\setcounter{theorem}{0}
\setcounter{lemma}{0}
\setcounter{proposition}{0}
\setcounter{corollary}{0}

\section{Gradient Computation Problem for the $\mathbb{E}_{\mathbf{x}}[L_\theta(\mathbf{x})]$ term}
\label{sec_grad_limitation}

The Neural Bayes parameterization contains the term $\mathbb{E}_{\mathbf{x}}[L_\theta(\mathbf{x})]$. Computing unbiased gradient through this term is in general difficult without the use of very large batch-sizes even though the quantity $\mathbb{E}_{\mathbf{x}}[L_\theta(\mathbf{x})]$ itself may have a good estimate using very few samples. For instance, consider the scalar function $\psi(t) = 1 + 0.01 \sin{\omega t}$. Consider the scenario when $\omega \rightarrow \infty$. The quantity $\mathbb{E}[\psi(t)]$ can be estimated very accurately using even one example. Further, $\mathbb{E}[\psi(t)]=1$, hence $\frac{\partial \mathbb{E}[\psi(t)]}{\partial t}=0$. However, when using a finite number of samples, the approximation of $\frac{\partial \mathbb{E}[\psi(t)]}{\partial t}$ can have a very high variance estimate due to improper cancelling of gradient terms from individual samples.

In the case of Neural Bayes-MIM we found that gradients through terms involving $\mathbb{E}_{\mathbf{x}}[L_\theta(\mathbf{x})]$ were 0. This allows us to estimate gradients for this objective reliably in the mini-batch setting. But in general it may be challenging to do so and solving objectives using Neural Bayes parameterization may require a customized work-around for each objective.

\section{Implementation Details of the Neural Bayes-MIM Objective}
\label{sec_lagnet_mim_applying}
We apply the Neural Bayes-MIM objective (Eq \ref{eq_MIMv2}) to all the hidden layers at different scales (using average pooling). We now discuss its implementation details. Consider the CNN architecture used in our experiments-- $C(200,3,1,0)-P(2,2,0,\text{max})-C(500,3,1,0)-C(700,3,1,0)-P(2,2,0,\text{max})-C(1000,3,1,0)$. Denote $\mathbf{h}^i$ ($i \in \{0,1,2,3 \}$) be the 4 hidden layer ReLU outputs after the 4 convolution layers. For input of size $32 \times 32 \times 3$, all these hidden states have height and width dimension in addition to channel dimension. For a mini-batch $B$, these hidden states are therefore 4 dimensional tensors. Let these 4 dimensions for the $i^{th}$ state be denoted by $ |B| \times C_i \times H_i \times W_i$, where the dimensions denote batch-size, number of channels, height and width.  Denote $\mathcal{S}$ to be the Softmax function applied along the channel dimension, and $\mathcal{P}$ to be $P(2,2,0,\text{avg})$. Further, denote $\mathbf{h}^i := \mathcal{P}(\mathbf{h}^{i-4})$ ($i \in \{4,5,6,7 \}$) as the scaled version of the original states computed by average pooling, and define numbers $C_i, H_i, W_i$ accordingly. Then the total Neural Bayes-MIM objective for this architecture is given by,
\begin{align}
    \min_{\theta} & -  \frac{1}{|B|}\sum_{\mathbf{x} \in B} \left[ \frac{1}{8}\sum_{i=0}^7  \frac{1}{H_i W_i}\sum_{h,w=1}^{H_i, W_i} \sum_{k=1}^{C_i} \mathcal{S}(\mathbf{h}^i_{k, h, w}(\mathbf{x})) \log \langle { \mathcal{S}(\mathbf{h}^i_{k, h, w}(\mathbf{x})) } +\epsilon \rangle \right] \nonumber\\ 
    & + (1+\alpha) \cdot \mathcal{R}_p(\theta) + \beta \cdot \mathcal{R}_c
\end{align}
where,
\begin{align}
    \mathcal{R}_p(\theta) &:= - \frac{1}{8}\sum_{i=0}^7  \frac{1}{H_i W_i}\sum_{h,w=1}^{H_i, W_i} \left[\sum_{k=1}^{C_i} \frac{1}{C_i} \log   \left[ \frac{1}{|B|}\sum_{\mathbf{x} \in B}\mathcal{S}(\mathbf{h}^i_{k, h, w}(\mathbf{x}))\right] +  \frac{C_i-1}{C_i}\log \left[1-  \frac{1}{|B|}\sum_{\mathbf{x} \in B}\mathcal{S}(\mathbf{h}^i_{k, h, w}(\mathbf{x})) \right] \right]
\end{align}
and,
\begin{align}
    \mathcal{R}_c =  \frac{1}{|B|}\sum_{\mathbf{x} \in B} \frac{ \lVert \mathcal{P}(\mathbf{h}^3_{k}(\mathbf{x})) - \mathcal{P}(\mathbf{h}^3_{k}(\mathbf{x} + \zeta \cdot \hat{\delta}) \rVert^2 }{\zeta^2} 
\end{align}
where $\hat{\delta} := \frac{\delta}{\lVert \delta \rVert_2}$ is a normalized noise vector computed independently for each sample $\mathbf{x}$ in the batch $B$ as,
\begin{align}
    \delta := \mathbf{X} \mathbf{v}.
\end{align}
Here $\mathbf{X} \in \mathbb{R}^{n \times B}$ is the matrix containing the batch of samples, and each dimension of $\mathbf{v} \in \mathbb{R}^B$ is sampled i.i.d. from a standard Gaussian. This computation ensures that the perturbation lies in the span of data. Finally $\zeta$ is the scale of normalized noise added to all samples in a batch. In our experiments, since we always normalize the datasets to have zero mean and unit variance across all dimensions, we sample $\zeta \sim \mathcal{N}(0,0.1^2)$. Note that for the architecture used, $\mathcal{P}(\mathbf{h}^3_{k}(\mathbf{x}))$ results in an output with height and width equal to 1, hence the output is effectively a 2D matrix of size $|B| \times C_3$. Finally, the gradient form this mini-batch is accumulated and averaged over multiple batches before updating the parameters for a more accurate estimate of gradients.

\section{Additional Analysis of Neural Bayes-MIM}

\subsection{Gradient Strength of Uniform Prior in Neural Bayes-MIM-v1 (Eq \ref{eq_decoupled_MIMv1}) vs Neural Bayes-MIM-v2 (\ref{eq_entropy_v2})}
\label{sec_uniform_prior}
As discussed in the main text, the term,
\begin{align}
    \mathcal{R}_p^{v1}(\theta) := \sum_{k=1}^K \mathbb{E}_{\mathbf{x}}[L_k(\mathbf{x})] \log \langle \mathbb{E}_{\mathbf{x}}[L_k(\mathbf{x})] \rangle
\end{align}
acts as a uniform prior encouraging the representations to be distributed. However, gradients are much stronger when $\mathbb{E}_{\mathbf{x}}[L_k(\mathbf{x})]$ approaches 1 for the alternative cross-entropy formulation,
\begin{align}
   \mathcal{R}_p^{v2}(\theta) := -\sum_{k=1}^K \frac{1}{K} \log  \mathbb{E}_{\mathbf{x}}[L_k(\mathbf{x})] +  \frac{K-1}{K}\log (1- \mathbb{E}_{\mathbf{x}}[L_k(\mathbf{x})] )
\end{align}
To see this, note that gradient for $\mathcal{R}_p^{v1}(\theta)$ is given by,
\begin{align}
    \frac{\partial \mathcal{R}_p^{v1}(\theta)}{\partial \theta} &=   \sum_{k=1}^K \frac{\partial  \mathbb{E}_{\mathbf{x}}[L_{\theta_k}(\mathbf{x})]}{\partial \theta} \log\mathbb{E}_{\mathbf{x}}[L_{\theta_k}(\mathbf{x})] - \sum_{k=1}^K \frac{\partial  \mathbb{E}_{\mathbf{x}}[L_{\theta_k}(\mathbf{x})]}{\partial \theta} \\
    &=   \sum_{k=1}^K \frac{\partial  \mathbb{E}_{\mathbf{x}}[L_{\theta_k}(\mathbf{x})]}{\partial \theta} \log\mathbb{E}_{\mathbf{x}}[L_{\theta_k}(\mathbf{x})] - \frac{\partial  \mathbb{E}_{\mathbf{x}}[\sum_{k=1}^K  L_{\theta_k}(\mathbf{x})]}{\partial \theta} \\
    &=   \sum_{k=1}^K \frac{\partial  \mathbb{E}_{\mathbf{x}}[L_{\theta_k}(\mathbf{x})]}{\partial \theta} \log\mathbb{E}_{\mathbf{x}}[L_{\theta_k}(\mathbf{x})]
\end{align}
where the last equality holds due to the linearity of expectation and because $\sum_{k=1}^K L_{\theta_k}(\mathbf{x})=1$ by design. On the other hand, gradients for $\mathcal{R}_p^{v2}(\theta)$ is given by,
\begin{align}
    \frac{\partial \mathcal{R}_p^{v2}(\theta)}{\partial \theta} &=    -\sum_{k=1}^K \frac{1}{K} \left( \frac{1}{\mathbb{E}_{\mathbf{x}}[L_k(\mathbf{x})]} - \frac{K-1}{1-\mathbb{E}_{\mathbf{x}}[L_k(\mathbf{x})]} \right) \frac{\partial  \mathbb{E}_{\mathbf{x}}[L_{\theta_k}(\mathbf{x})]}{\partial \theta}
\end{align}
When the representation being learned is such that the marginal $p(z)$ peaks along a single state $k$, i.e., $\mathbb{E}_{\mathbf{x}}[L_k(\mathbf{x})] \rightarrow 1$ (making the representation degenerate), the gradient for the $k^{th}$ term for v1 is given by,
\begin{align}
    \frac{\partial  \mathbb{E}_{\mathbf{x}}[L_{\theta_k}(\mathbf{x})]}{\partial \theta} \log\mathbb{E}_{\mathbf{x}}[L_{\theta_k}(\mathbf{x})] \approx 0
\end{align}
while that for v2 is given by,
\begin{align}
     -\frac{1}{K} \left( \frac{1}{\mathbb{E}_{\mathbf{x}}[L_k(\mathbf{x})]} - \frac{K-1}{1-\mathbb{E}_{\mathbf{x}}[L_k(\mathbf{x})]} \right) \frac{\partial  \mathbb{E}_{\mathbf{x}}[L_{\theta_k}(\mathbf{x})]}{\partial \theta} \approx \lim_{c \rightarrow 0} \frac{1}{c} \cdot \frac{\partial  \mathbb{E}_{\mathbf{x}}[L_{\theta_k}(\mathbf{x})]}{\partial \theta}
\end{align}
whose magnitude approaches infinity as $\mathbb{E}_{\mathbf{x}}[L_k(\mathbf{x})] \rightarrow 1$. Thus $\mathcal{R}_p^{v2}(\theta)$ is beneficial in terms of gradient strength.

\subsection{Empirical Comparison between Neural Bayes-MIM-v1 (Eq \ref{eq_decoupled_MIMv1}) and Neural Bayes-MIM-v2 (\ref{eq_entropy_v2})}
\label{sec_v1_vs_v2}

\begin{figure}
\centering
\includegraphics[trim={0cm 1.3cm 0 1.2cm},clip, width=0.5\linewidth]{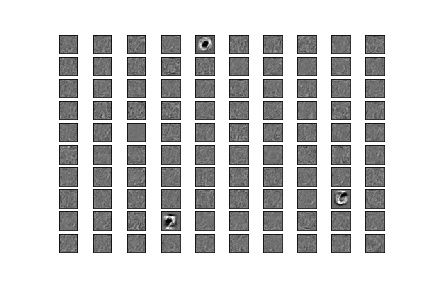} 
\caption{MNIST filters learned using Neural Bayes-MIM-v1 objective (Eq \ref{eq_decoupled_MIMv1}) using the configuration $\alpha=4$, $\beta=4$. Majority of filters are dead. For comparison with filters from Neural Bayes-MIM-v2, see figure \ref{fig:mnist_filters_full} (bottom).}
\label{fig:mnist_filters_v1}
 
\end{figure}

\begin{figure}
\centering
 
\subfloat[$\alpha=0$, $\beta=4$]{
	\includegraphics[trim={0cm 1.3cm 0 1.2cm},clip, width=0.5\linewidth]{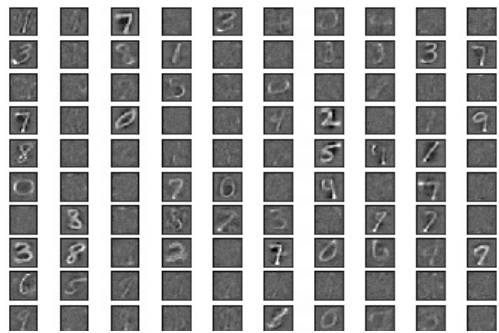} } 
 
\subfloat[$\alpha=4$, $\beta=0$]{
	\includegraphics[trim={0cm 1.3cm 0 1.2cm},clip, width=0.5\linewidth]{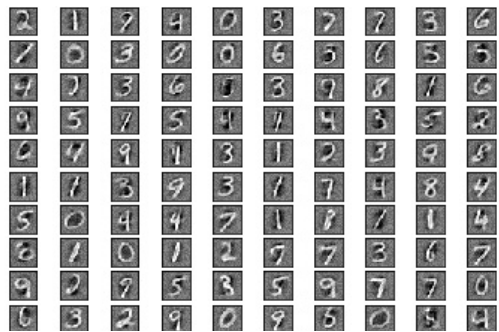} } 
 
\subfloat[$\alpha=4$, $\beta=4$]{
	\includegraphics[trim={0cm 1.3cm 0 1.2cm},clip, width=0.5\linewidth]{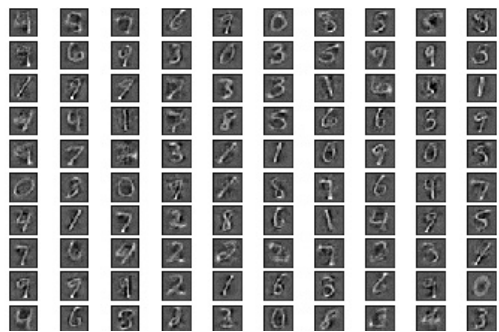} } 
 
\caption{MNIST filters learned using Neural Bayes-MIM objective Eq \ref{eq_MIMv2}. Majority of filters are dead when regularization coefficient $\alpha=0$. Filters memorize input samples when regularization coefficient $\beta=0$. Using both regularization terms results in filters that mainly capture parts of inputs which are good for distributed representation.}
\label{fig:mnist_filters_full}
 
\end{figure}

To empirically understand the difference in behavior of Neural Bayes-MIM objective v1 vs v2, we first plot the filters learned by the v1 objective and compare it with those learned by the v2 objective. The filters learned by the v1 objective are shown in figure \ref{fig:mnist_filters_v1} using the configuration $\alpha=4$, $\beta=4$. It can be seen that most filters are dead. We tried other configurations as well without any change in the outcome. Since the v1 and v2 objective differ only in the formulation of the uniform prior regularization, as explained in the previous section, we believe that v1 leads to dead filters because of weak gradients from its regularization term.

\begin{figure*}
\centering
  \includegraphics[trim={0cm 1.7cm 0 5cm},clip, width=1\linewidth]{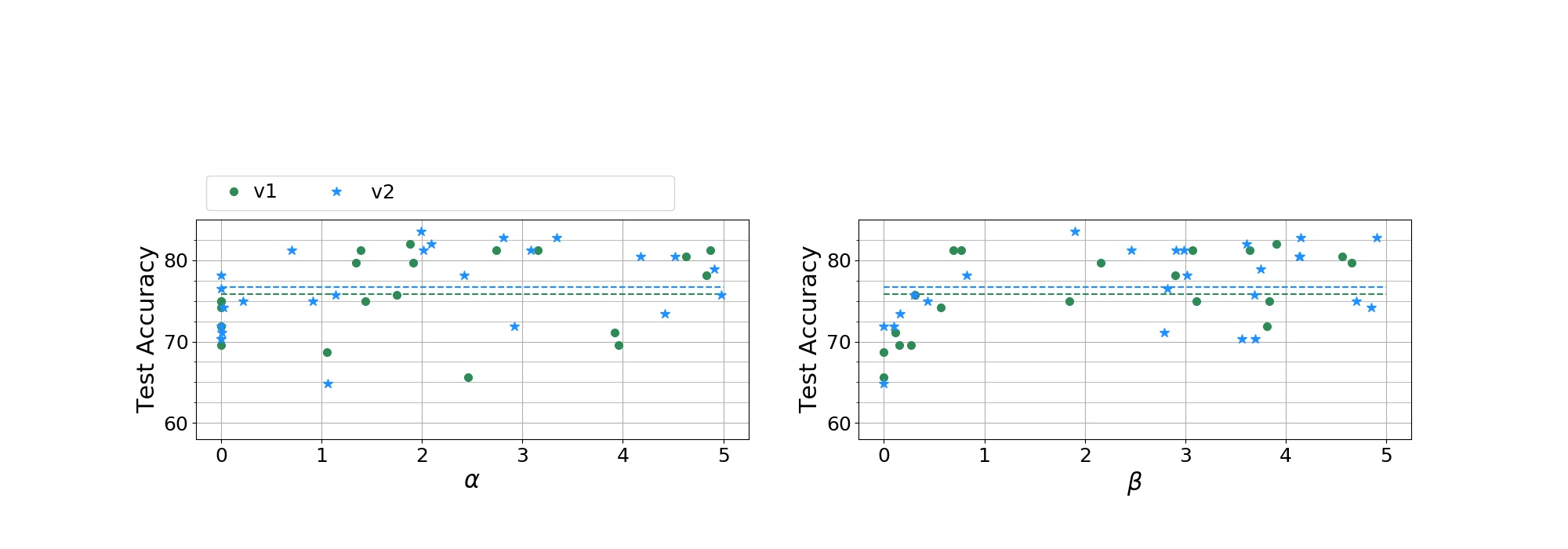}
  \caption{Performance of Neural Bayes-MIM-v1 vs Neural Bayes-MIM-v2. A CNN encoder is trained using Neural Bayes-MIM with different configurations-- hyper-parameters $\alpha$, $\beta$ and scaling of states on which the objective is applied. A one hidden layer classifier (with 200 units) is then trained using labels on these frozen features to get the final test accuracy.Green and blue dotted lines are the average of all the green and blue points respectively. Neural Bayes-MIM-v2 has better average and best performance compared with Neural Bayes-MIM-v1.}
  \label{fig:v1_vs_v2_ablation}
\end{figure*}

In the second set of experiments, we train many models using Neural Bayes-MIM-v1 and Neural Bayes-MIM-v2 objectives separately with different hyper-parameter configurations similar to the setting of figure \ref{fig:c10_3ablation}. The performance scatter plot is shown in figure \ref{fig:v1_vs_v2_ablation}. We find that Neural Bayes-MIM-v2 has better average and best performance compared with Neural Bayes-MIM-v1.

\section{Proof of Lemma \ref{lemma_reparam_trick}}

\begin{lemma}
Let $p(\mathbf{x}|z=k)$ and $p(z)$ be any conditional and marginal distribution defined for continuous random variable $\mathbf{x}$ and discrete random variable $z$. If $\mathbb{E}_{\mathbf{x}\sim p(\mathbf{x})}[L_k(\mathbf{x})] \neq 0$ $\forall k \in [K]$, then there exists a non-parametric function $L(\mathbf{x}): \mathbb{R}^n \rightarrow \mathbb{R}^{+^{K}}$ for any given input $\mathbf{x} \in \mathbb{R}^n$ with the property $\sum_{k=1}^K L_k(\mathbf{x})=1$ $\forall \mathbf{x}$ such that,
\begin{align}
    p(\mathbf{x}|z=k) = \frac{L_k(\mathbf{x}) \cdot p(\mathbf{x})}{\mathbb{E}_{\mathbf{x}\sim p(\mathbf{x})}[L_k(\mathbf{x})]}, \mspace{20mu} p(z=k) = \mathbb{E}_{\mathbf{x}}[L_k(\mathbf{x})], \mspace{20mu} p(z=k|\mathbf{x})= L_k(\mathbf{x})
\end{align}
and this parameterization is consistent.

\textbf{Proof}: First we show the existence proof. Notice that there exists a non-parametric function $g_k(\mathbf{x}) := \frac{p(\mathbf{x}|z=k)}{p(\mathbf{x})}$ $\forall \mathbf{x} \in supp(p(\mathbf{x}))$. Denote $G_k(\mathbf{x}) = p(z=k) g_k(\mathbf{x})$. Then,
\begin{align}
    \mathbb{E}_{\mathbf{x}}[G_k(\mathbf{x})] = \mathbb{E}_{\mathbf{x}}[p(z=k) g_k(\mathbf{x})] = p(z=k)
\end{align}
and, 
\begin{align}
    \frac{G_k(\mathbf{x})}{\mathbb{E}_{\mathbf{x}}[G_k(\mathbf{x})]} &= \frac{p(z=k) g_k(\mathbf{x})}{p(z=k)} = \frac{p(\mathbf{x}|z=k)}{p(\mathbf{x})}
\end{align}
Thus $L_k := G_k$ works. To verify that this parameterization is consistent, note that for any $k$,
\begin{align}
    \int_{\mathbf{x}} p(\mathbf{x}|z=k) &= \int_{\mathbf{x}} \frac{L_k(\mathbf{x}) \cdot p(\mathbf{x})}{\mathbb{E}_{\mathbf{x}\sim p(\mathbf{x})}[L_k(\mathbf{x})]}=1
\end{align}
where we use the condition $\mathbb{E}_{\mathbf{x}\sim p(\mathbf{x})}[L_k(\mathbf{x})] \neq 0$ $\forall k \in [K]$. Secondly, we note that,
\begin{align}
    \sum_{k=1}^K p(\mathbf{x}|z=k)\cdot p(z=k) &=
    \sum_{k=1}^K \frac{L_k(\mathbf{x}) \cdot p(\mathbf{x})}{\mathbb{E}_{\mathbf{x}}[L_k(\mathbf{x})]} \cdot \mathbb{E}_{\mathbf{x}}[L_k(\mathbf{x})]\\
    &= \sum_{k=1}^K L_k(\mathbf{x}) \cdot p(\mathbf{x})\\
    &= p(\mathbf{x})
\end{align}
where the last equality is due to the conditions $\sum_{k=1}^K L_k(\mathbf{x})=1$ $\forall \mathbf{x}$. Thirdly,
\begin{align}
    \sum_{k=1}^K p(z=k) &= \sum_{k=1}^K \mathbb{E}_{\mathbf{x}}[L_k(\mathbf{x})]\\
    &=  \mathbb{E}_{\mathbf{x}}[ \sum_{k=1}^K L_k(\mathbf{x})]\\
    &=1 \mspace{70mu} \nonumber
\end{align}

Finally, we have from Bayes' rule:
\begin{align}
\label{eq_L_posterior}
    p(z=k|\mathbf{x}) &= \frac{p(\mathbf{x}|z=k) \cdot p(z=k)}{p(\mathbf{x})}\\
    &= \frac{L_k(\mathbf{x}) \cdot p(\mathbf{x})}{\mathbb{E}_{\mathbf{x}\sim p(\mathbf{x})}[L_k(\mathbf{x})]} \cdot \frac{\mathbb{E}_{\mathbf{x}\sim p(\mathbf{x})}[L_k(\mathbf{x})]}{p(\mathbf{x})}\\
    &= L_k(\mathbf{x})
\end{align}
where the second equality holds because of the existence and consistency proofs of $p(\mathbf{x}|z=k) := \frac{L_k(\mathbf{x}) \cdot p(\mathbf{x})}{\mathbb{E}_{\mathbf{x}\sim p(\mathbf{x})}[L_k(\mathbf{x})]}$ and $ p(z=k) := \mathbb{E}_{\mathbf{x}}[L_k(\mathbf{x})]$ shown above.
$\square$
\end{lemma}

\section{Proofs for Neural Bayes-MIM}

\begin{proposition} (Neural Bayes-MIM-v1)
(proposition 1 in main text)
Let $L(\mathbf{x}): \mathbb{R}^n \rightarrow \mathbb{R}^{+^{K}}$ be a non-parametric function for any given input $\mathbf{x} \in \mathbb{R}^n$ with the property $\sum_{i=1}^K L_k(\mathbf{x})=1$ $\forall \mathbf{x}$. Consider the following objective,
\begin{align}
    L^* = \arg\max_{L} \mathbb{E}_{\mathbf{x}} \left[ \sum_{k=1}^K L_k(\mathbf{x}) \log \frac{L_k(\mathbf{x})}{\mathbb{E}_{\mathbf{x}}[L_k(\mathbf{x})]} \right]
\end{align}
Then $L^*_k(\mathbf{x}) = p(z^*=k|\mathbf{x})$, where $ z^* \in \arg\max_{z} MI(\mathbf{x},z)$.

\textbf{Proof}: Using the Neural Bayes parameterization in lemma \ref{lemma_reparam_trick}, we have,
\begin{align}
    &MI(\mathbf{x},z) = \int_{\mathbf{x}} \sum_{k=1}^K p(\mathbf{x},z=k) \log \frac{p(\mathbf{x},z=k)}{p(\mathbf{x}) p(z=k)}\\
    &= \int_{\mathbf{x}} \sum_{k=1}^K p(z=k|\mathbf{x}) p(\mathbf{x}) \log \frac{p(z=k|\mathbf{x}) }{p(z=k)}\\
    &= \int_{\mathbf{x}} \sum_{k=1}^K {L_k(\mathbf{x}) \cdot p(\mathbf{x})} \cdot  \log \frac{L_k(\mathbf{x}) }{\mathbb{E}_{\mathbf{x}}[L_k(\mathbf{x})]}\\
    &= \mathbb{E}_{\mathbf{x} \sim p(\mathbf{x})} \left[ \sum_{k=1}^K L_k(\mathbf{x}) \log \frac{L_k(\mathbf{x})}{\mathbb{E}_{\mathbf{x}}[L_k(\mathbf{x})]} \right]
\end{align}
Therefore the two objectives are equivalent and we have a closed form estimate of mutual information. Given $z^*$ is a maximizer of $MI(\mathbf{x},z)$, since $L$ is a non-parametric function, there exists $L^*$ such that $p(z^*=k|\mathbf{x}) = {L_k^*(\mathbf{x}) }$ due to lemma \ref{lemma_reparam_trick}. $\square$

\end{proposition}

\begin{theorem}
 (Theorem 1 in main text) Denote,
\begin{align}
    J(\theta) = -\mathbb{E}_{\mathbf{x}} \left[ \sum_{k=1}^K L_{\theta_k}(\mathbf{x}) \log \frac{L_{\theta_k}(\mathbf{x})}{\mathbb{E}_{\mathbf{x}}[L_{\theta_k}(\mathbf{x})]} \right]
\end{align}
\begin{align}
    \hat{J}(\theta) =  -\mathbb{E}_{\mathbf{x}} \left[ \sum_{k=1}^K L_{\theta_k}(\mathbf{x}) \log \langle  \frac{L_{\theta_k}(\mathbf{x})}{\mathbb{E}_{\mathbf{x}}[L_{\theta_k}(\mathbf{x})]} \rangle\right]
\end{align}
where $\langle .\rangle$ denotes gradients are not computed through the argument. Then $\frac{\partial J(\theta)}{\partial \theta} = \frac{\partial \hat{J}(\theta)}{\partial \theta}$.

\textbf{Proof}: We note that,
\begin{align}
    J(\theta) &= -\mathbb{E}_{\mathbf{x}} \left[ \sum_{k=1}^K L_{\theta_k}(\mathbf{x}) \log \frac{L_{\theta_k}(\mathbf{x})}{\mathbb{E}_{\mathbf{x}}[L_{\theta_k}(\mathbf{x})]} \right]\\
    &= -\mathbb{E}_{\mathbf{x}} \left[ \sum_{k=1}^K L_{\theta_k}(\mathbf{x}) \log L_{\theta_k}(\mathbf{x}) \right] + \mathbb{E}_{\mathbf{x}} \left[ \sum_{k=1}^K L_{\theta_k}(\mathbf{x}) \log \mathbb{E}_{\mathbf{x}}[L_{\theta_k}(\mathbf{x})] \right]
\end{align}
Denote the first term by $T_1$. Then due to chain rule,
\begin{align}
    -\frac{\partial T_1}{\partial \theta} &=  \mathbb{E}_{\mathbf{x}} \left[ \sum_{k=1}^K \frac{\partial  L_{\theta_k}(\mathbf{x})}{\partial \theta} \log L_{\theta_k}(\mathbf{x}) \right] - \mathbb{E}_{\mathbf{x}} \left[ \sum_{k=1}^K \frac{L_{\theta_k}(\mathbf{x})}{L_{\theta_k}(\mathbf{x})} \cdot \frac{\partial  L_{\theta_k}(\mathbf{x})}{\partial \theta}  \right]\\
    &=  \mathbb{E}_{\mathbf{x}} \left[ \sum_{k=1}^K \frac{\partial  L_{\theta_k}(\mathbf{x})}{\partial \theta} \log L_{\theta_k}(\mathbf{x}) \right] - \mathbb{E}_{\mathbf{x}} \left[ \sum_{k=1}^K \frac{\partial  L_{\theta_k}(\mathbf{x})}{\partial \theta}  \right]\\
    &=  \mathbb{E}_{\mathbf{x}} \left[ \sum_{k=1}^K \frac{\partial  L_{\theta_k}(\mathbf{x})}{\partial \theta} \log L_{\theta_k}(\mathbf{x}) \right] - \mathbb{E}_{\mathbf{x}} \left[ \frac{\partial \sum_{k=1}^K  L_{\theta_k}(\mathbf{x})}{\partial \theta}  \right]\\
    &=  \mathbb{E}_{\mathbf{x}} \left[ \sum_{k=1}^K \frac{\partial  L_{\theta_k}(\mathbf{x})}{\partial \theta} \log L_{\theta_k}(\mathbf{x}) \right]
\end{align}
where the last equality holds due to the linearity of expectation and because $\sum_{k=1}^K L_{\theta_k}(\mathbf{x})=1$ by design. Now denote the second term by $T_2$. Then due to chain rule,
\begin{align}
    -\frac{\partial T_2}{\partial \theta} &=  -\mathbb{E}_{\mathbf{x}} \left[ \sum_{k=1}^K \frac{\partial  L_{\theta_k}(\mathbf{x})}{\partial \theta} \log \mathbb{E}_{\mathbf{x}}[L_{\theta_k}(\mathbf{x})] \right] - \mathbb{E}_{\mathbf{x}} \left[ \sum_{k=1}^K \frac{L_{\theta_k}(\mathbf{x})}{\mathbb{E}_{\mathbf{x}}[L_{\theta_k}(\mathbf{x})]} \cdot \frac{\partial  \mathbb{E}_{\mathbf{x}}[L_{\theta_k}(\mathbf{x})]}{\partial \theta}  \right]\\
    &=  -\mathbb{E}_{\mathbf{x}} \left[ \sum_{k=1}^K \frac{\partial  L_{\theta_k}(\mathbf{x})}{\partial \theta} \log \mathbb{E}_{\mathbf{x}}[L_{\theta_k}(\mathbf{x})] \right] -   \sum_{k=1}^K \frac{ \mathbb{E}_{\mathbf{x}}[L_{\theta_k}(\mathbf{x})]}{\mathbb{E}_{\mathbf{x}}[L_{\theta_k}(\mathbf{x})]} \cdot \frac{\partial  \mathbb{E}_{\mathbf{x}}[L_{\theta_k}(\mathbf{x})]}{\partial \theta} \\
    &= -\mathbb{E}_{\mathbf{x}} \left[ \sum_{k=1}^K \frac{\partial  L_{\theta_k}(\mathbf{x})}{\partial \theta} \log \mathbb{E}_{\mathbf{x}}[L_{\theta_k}(\mathbf{x})] \right] -  \sum_{k=1}^K \mathbb{E}_{\mathbf{x}}\left[\frac{\partial  L_{\theta_k}(\mathbf{x})}{\partial \theta}\right] \\
    &= -\mathbb{E}_{\mathbf{x}} \left[ \sum_{k=1}^K \frac{\partial  L_{\theta_k}(\mathbf{x})}{\partial \theta} \log \mathbb{E}_{\mathbf{x}}[L_{\theta_k}(\mathbf{x})] \right] -  \mathbb{E}_{\mathbf{x}}\left[\frac{\partial  \sum_{k=1}^K L_{\theta_k}(\mathbf{x})}{\partial \theta}\right]  \\
    &= -\mathbb{E}_{\mathbf{x}} \left[ \sum_{k=1}^K \frac{\partial  L_{\theta_k}(\mathbf{x})}{\partial \theta} \log \mathbb{E}_{\mathbf{x}}[L_{\theta_k}(\mathbf{x})] \right]
\end{align}
where once again the last equality holds due to the linearity of expectation and because $\sum_{k=1}^K L_{\theta_k}(\mathbf{x})=1$ by design. Thus the gradient for $J$ is given by,
\begin{align}
    \frac{\partial J(\theta)}{\partial \theta} = -\mathbb{E}_{\mathbf{x}} \left[ \sum_{k=1}^K \frac{\partial  L_{\theta_k}(\mathbf{x})}{\partial \theta} \log \frac{L_{\theta_k}(\mathbf{x})}{\mathbb{E}_{\mathbf{x}}[L_{\theta_k}(\mathbf{x})]}  \right]
\end{align}
which is the same as $\frac{\partial \hat{J}(\theta)}{\partial \theta}$. This concludes the proof. $\square$
\end{theorem}

\section{Proofs for Neural Bayes-DML (binary case)}

\begin{proposition}
(Neural Bayes-DML, proposition 2 in main text)
Let $L(\mathbf{x}): \mathbb{R}^n \rightarrow [0,1]$ be a non-parametric function for any given input $\mathbf{x} \in \mathbb{R}^n$, and let $J$ be the Jensen-Shannon divergence. Define scalars $f_1(\mathbf{x}) := 
\frac{L(\mathbf{x})}{\mathbb{E}_{\mathbf{x}}[L(\mathbf{x})]}$ and $f_0(\mathbf{x}) := 
\frac{1-L(\mathbf{x})}{1-\mathbb{E}_{\mathbf{x}}[L(\mathbf{x})]}$. Then the objective in Eq. (\ref{eq_primary_objective}) is equivalent to,
\begin{align}
    \max_{L} &\frac{1}{2} \cdot \mathbb{E}_{\mathbf{x}} \left[ f_1(\mathbf{x}) \cdot \log \left( \frac{f_1(\mathbf{x})}{f_1(\mathbf{x}) + f_0(\mathbf{x})} \right) \right] + \frac{1}{2} \cdot \mathbb{E}_{\mathbf{x}} \left[f_0(\mathbf{x}) \cdot \log \left( \frac{f_0(\mathbf{x})}{f_1(\mathbf{x}) + f_0(\mathbf{x})} \right) \right] + \log 2  \\
    & s.t. \mspace{20mu}  \mathbb{E}_{\mathbf{x}}[L(\mathbf{x})] \notin \{0,1\}
\end{align}

\textbf{Proof}: Using the Neural Bayes parameterization from lemma \ref{lemma_reparam_trick} for binary case, we set,
\begin{align}
\label{eq_binary_parameterization}
    q_1(\mathbf{x}) := \frac{L(\mathbf{x}) \cdot p(\mathbf{x})}{\mathbb{E}_{\mathbf{x}\sim p(\mathbf{x})}[L(\mathbf{x})]} \mspace{20mu} q_0(\mathbf{x}) := \frac{(1 - L(\mathbf{x})) \cdot p(\mathbf{x})}{1 - \mathbb{E}_{\mathbf{x}\sim p(\mathbf{x})}[L(\mathbf{x})]}
\end{align}
These parameterizations therefore automatically satisfy the constraints in Eq. (\ref{eq_primary_objective}). Finally, using the definition of JS divergence, the maximization problem in Eq. (\ref{eq_primary_objective}) can be written as,
\begin{align}
    \max_{q_0,q_1} &\frac{1}{2}\cdot \int_{\mathbf{x}} q_1(\mathbf{x}) \log \frac{q_1(\mathbf{x})}{0.5 \cdot (q_0(\mathbf{x}) + q_1(\mathbf{x}))}+ q_0(\mathbf{x}) \log \frac{q_0(\mathbf{x})}{0.5 \cdot (q_0(\mathbf{x}) + q_1(\mathbf{x}))}
\end{align}
Substituting $q_0$ and $q_1$ with their respective parameterizations and using the definitions of $f_0(\mathbf{x})$ and $f_1(\mathbf{x})$ completes the proof. $\square$
\end{proposition}

\begin{theorem} (optimality, Theorem 2 in main text)
Let $p(\mathbf{x})$ be a probability density function over $\mathbb{R}^n$ whose support is the union of two non-empty connected sets (definition \ref{def_conn_set}) $S_1$ and $S_2$ that are disjoint, i.e. $S_1 \cap S_2 \ = \varnothing$. Let $L(\mathbf{x}) \in [0,1]$ belong to the class of continuous functions which is learned by solving the objective in Eq. (\ref{eq_binary_obj_theorem}). Then the objective in Eq. (\ref{eq_binary_obj_theorem}) is maximized if and only if one of the following is true:
\begin{align}
\label{eq_optimality}
L(\mathbf{x}) =
    \begin{cases}
    0 & \forall \mathbf{x} \in S_1 \\
    1 & \forall \mathbf{x} \in S_2
    \end{cases} \mspace{30mu} \text{or} \mspace{30mu}  L(\mathbf{x}) =
    \begin{cases}
    1 & \forall \mathbf{x} \in S_1 \\
    0 & \forall \mathbf{x} \in S_2
    \end{cases}
\end{align}

\textbf{Proof}: The two cases exist in the theorem due to symmetry. Recall the definition of
$f_0(\mathbf{x})$ and $f_1(\mathbf{x})$ in Eq. (\ref{eq_binary_obj_theorem}),
\begin{align}
f_1(\mathbf{x}) := 
\frac{L(\mathbf{x})}{\mathbb{E}_{\mathbf{x}}[L(\mathbf{x})]} \mspace{40mu} f_0(\mathbf{x}) := 
\frac{1-L(\mathbf{x})}{1-\mathbb{E}_{\mathbf{x}}[L(\mathbf{x})]}
\end{align}
where $L(\mathbf{x}) \in [0,1]$ and for a feasible $L(\mathbf{x})$, and therefore $\pi := \mathbb{E}_{\mathbf{x}}[L(\mathbf{x})] \in (0,1)$ due to the conditions specified in this theorem. Thus $f_1(\mathbf{x}) \in [0,\frac{1}{\pi}]$ and $f_0(\mathbf{x}) \in [0,\frac{1}{1-\pi}]$. By design, the terms $\log \left( \frac{f_1(\mathbf{x})}{f_1(\mathbf{x}) + f_0(\mathbf{x})} \right)$ and $\log \left( \frac{f_0(\mathbf{x})}{f_1(\mathbf{x}) + f_0(\mathbf{x})} \right)$ are non-positive. Thus, for any $\mathbf{x} \in S_1 \cup S_2$,
\begin{align}
    F(\mathbf{x}) = f_1(\mathbf{x}) \cdot \log \left( \frac{f_1(\mathbf{x})}{f_1(\mathbf{x}) + f_0(\mathbf{x})} \right) + f_0(\mathbf{x}) \cdot \log \left( \frac{f_0(\mathbf{x})}{f_1(\mathbf{x}) + f_0(\mathbf{x})} \right)
\end{align}
is maximized only when $L(\mathbf{x}) = 0$ or $L(\mathbf{x}) = 1$ leading to $F(\mathbf{x})=0$. Therefore, the objective in Eq. (\ref{eq_binary_obj_theorem}) is maximized by setting $L(\mathbf{x}) = 0$ or $L(\mathbf{x}) = 1$ $\forall \mathbf{x} \in S_1 \cup S_2$.

Finally, since $L(\mathbf{x})$ is a continuous function, $\nexists \mathbf{x}_1, \mathbf{x}_2 \in S_1$ such that $L(\mathbf{x}_1) = 0$ and $L(\mathbf{x}_2) = 1$. We prove this by contradiction. Suppose there exists a pair $(\mathbf{x}_1, \mathbf{x}_2)$ of this kind. Then along any path connecting $\mathbf{x}_1$ and $\mathbf{x}_2$ within $S_1$, there must exist a point where $L(\mathbf{x})$ is not continuous since $L(\mathbf{x}) = 0$ or $L(\mathbf{x}) = 1$ $\forall \mathbf{x} \in S_1 \cup S_2$ to satisfy the maximization condition. This is a contradiction. By symmetry, the same argument can be proved for $\mathbf{x}_1, \mathbf{x}_2 \in S_2$. Therefore one of the two cases mentioned in the theorem must be the optimal solution for $L(\mathbf{x})$ in Eq. (\ref{eq_binary_obj_theorem}). Thus we have proved the claim. $\square$
\end{theorem}

\section{Neural Bayes-DML: Extension to Multiple Partitions}
\label{sec_lagnet_dml_extension}
In order to extend our proposal to multiple partitions (say $K$), the idea is to find conditional distribution $q_i$ ($i \in [K]$) corresponding to each of the $K$ partitions such the divergence between conditional distribution of every partition and the conditional distribution of the combined remaining partitions is maximized. Specifically, we propose the following primary objective,

\begin{align}
\label{eq_multi_primary_objective}
    &\max_{q_k \atop \pi_k\neq 0, \forall k \in [K]} \frac{1}{K}\sum_{k=1}^K J(q_k(\mathbf{x}) || \bar{q}_k(\mathbf{x})) \mspace{20mu} \text{s.t.}\\
    & \int_{\mathbf{x}} q_k(\mathbf{x})=1 \mspace{20mu} \forall k \in [K]\\
    &  \sum_{k=1}^K q_k(\mathbf{x})\cdot \pi_k = p(\mathbf{x})\\
    & \sum_{k=1}^K \pi_k = 1
\end{align}
where $\bar{q}_k(\mathbf{x})$ is the conditional distribution corresponding to the full data distribution excluding the partition defined by $q_k(\mathbf{x})$. Formally,
\begin{align}
    \bar{q}_k(\mathbf{x}) := \frac{p(\mathbf{x}) - q_{k}(\mathbf{x})\cdot \pi_k}{1 - \pi_k}
\end{align}

Then the theorem below shows an equivalent way of solving the above objective.
\begin{theorem}
\label{main_theorem_multi}
Let $L(\mathbf{x}): \mathbb{R}^n \rightarrow \mathbb{R}^{+^{K}}$ be a non-parametric function for any given input $\mathbf{x} \in \mathbb{R}^n$ with the property $\sum_{k=1}^K L_k(\mathbf{x})=1$ $\forall \mathbf{x}$, and let $J$ be the Jensen-Shannon divergence.
Define scalars $f_k(\mathbf{x}) := 
\frac{L_k(\mathbf{x})}{\mathbb{E}_{\mathbf{x}}[L_k(\mathbf{x})]}$ and $\bar{f}_k(\mathbf{x}) := 
\frac{1-L_k(\mathbf{x})}{1-\mathbb{E}_{\mathbf{x}}[L_k(\mathbf{x})]}$. Then the objective in Eq. (\ref{eq_multi_primary_objective}) is equivalent to,
\begin{align}
    &\max_{L_k \forall i \in [K]} \frac{1}{2} \cdot \mathbb{E}_{\mathbf{x}} \left[ \sum_{k=1}^K f_k(\mathbf{x}) \cdot \log \left( \frac{f_k(\mathbf{x})}{{f}_k(\mathbf{x}) + \bar{f}_k(\mathbf{x})} \right) + \bar{f}_k(\mathbf{x}) \cdot \log \left( \frac{\bar{f}_k(\mathbf{x})}{\bar{f}_k(\mathbf{x}) + f_k(\mathbf{x})} \right) \right] + \log 2\\
    &  \text{s.t.} \mspace{30mu} \mathbb{E}_{\mathbf{x}}[L_k(\mathbf{x})] = \pi_k \nonumber
\end{align}
Here $L_k(\mathbf{x})$ denotes the $k^{th}$ unit of $L(\mathbf{x})$.

\textbf{Proof}:  Similar to theorem \ref{main_theorem}, the main idea is to parameterize $q_k$ and $\bar{q}_k$ as follows,
\begin{align}
    q_k(\mathbf{x}) := \frac{L_k(\mathbf{x}) \cdot p(\mathbf{x})}{\mathbb{E}_{\mathbf{x}\sim p(\mathbf{x})}[L_k(\mathbf{x})]} \mspace{20mu} \bar{q}_k(\mathbf{x}) := \frac{(1 - L_k(\mathbf{x})) \cdot p(\mathbf{x})}{1 - \mathbb{E}_{\mathbf{x}\sim p(\mathbf{x})}[L_k(\mathbf{x})]}
\end{align}
To verify that these parameterizations are valid, note that,
\begin{align}
    \int_{\mathbf{x}} q_k(\mathbf{x}) &= \int_{\mathbf{x}} \frac{L_k(\mathbf{x}) \cdot p(\mathbf{x})}{\mathbb{E}_{\mathbf{x}\sim p(\mathbf{x})}[L_k(\mathbf{x})]}=1
\end{align}
Similarly, $ \int_{\mathbf{x}} \bar{q}_k(\mathbf{x})=1$. To verify that the second constraint is satisfied, we use the above parameterization and substitute $\mathbb{E}_{\mathbf{x}}[L_k(\mathbf{x})] = \pi_k$ and get,
\begin{align}
    \sum_{k=1}^K \frac{L_k(\mathbf{x}) \cdot p(\mathbf{x})}{\pi_k}\cdot \pi_k &= p(\mathbf{x}) \cdot \left(\sum_{k=1}^K L_k(\mathbf{x}) \right)\\
    &= p(\mathbf{x})
\end{align}
where the last equality uses the definition of $L(\mathbf{x})$. Also notice that each $\pi_k \in [0,1]$ and thus $\mathbb{E}_{\mathbf{x}}[L_k(\mathbf{x})] = \pi_k$ is feasible for any arbitrary distribution $q_k(\mathbf{x})$ when $L_k(\mathbf{x})\geq 0$.

Finally, using the proposed parameterization we have,
\begin{align}
    \bar{q}_k(\mathbf{x}) &= \frac{p(\mathbf{x}) - q_{i}(\mathbf{x})\cdot \pi_k}{1 - \pi_k}\\
    &= p(\mathbf{x}) \cdot \frac{1 - \frac{L^{i}(\mathbf{x})}{\mathbb{E}_{\mathbf{x}}[L_k(\mathbf{x})]}\cdot \pi_k}{1 - \pi_k}\\
    &= p(\mathbf{x}) \cdot \frac{1 - {L^{i}(\mathbf{x})}}{1 - \mathbb{E}_{\mathbf{x}}[L_k(\mathbf{x})]}\\
    &= \bar{f}_k(\mathbf{x}) \cdot p(\mathbf{x})
\end{align}
where we have used the fact that $\mathbb{E}_{\mathbf{x}}[L_k(\mathbf{x})] = \pi_k$. Using the definition of JS divergence, the max problem in Eq. (\ref{eq_multi_primary_objective}) can be written as,
\begin{align}
    &\max_{L_k \forall i \in [K]} \frac{1}{2} \cdot \sum_{k=1}^K  \int_\mathbf{x} q_k(\mathbf{x}) \cdot \log \left( \frac{q_k(\mathbf{x})}{0.5\cdot ({q}_k(\mathbf{x}) + \bar{q}_k(\mathbf{x}))} \right) + \bar{q}_k(\mathbf{x}) \cdot \log \left( \frac{\bar{q}_k(\mathbf{x})}{0.5\cdot (\bar{q}_k(\mathbf{x}) + q_k(\mathbf{x}))} \right)
\end{align}
Substituting $q_k$ and $\bar{q}_k$ with their respective parameterizations and using the definitions of $f_k(\mathbf{x})$ and $\bar{f}_k(\mathbf{x})$ completes the proof. $\square$
\end{theorem}
In terms of implementation, we propose to simply have $K$ output units in the label generating network $L_\theta$ while sharing the rest of the network. Also, we use Softmax activation at the output layer to satisfy the properties of $L$ specified in the above theorem.

\section{Additional Neural Bayes-DML Experiments}
\label{sec_lagnet_dml_mnist}

We run an experiment on MNIST. We randomly split the training set into $90\%-10\%$ training-validation set In this experiment, we train a CNN with the following architecture: $C(100,3,1,0)-P(2,2,0,\text{max})-C(100,3,1,0)-C(200,3,1,0)-P(2,2,0,\text{max})-C(500,3,1,0)-P(.,.,.,\text{avg})-FC(10)$. Here $P(.,.,.,\text{avg})$ denotes the entire spatial field is average pooled to result in $1\times 1$ height-width, and $FC(10)$ denotes a fully connected layer with output dimension 10. Finally, Softmax is applied at the output and the network is trained using the Neural Bayes-DML objective. We optimize the objective using Adam with learning rate 0.001, batch size 5000, 0 weight decay for 100 epochs (other Adam hyper-parameters are kept standard). We use $\beta = 1$ for the smoothness regularization coefficient. Once this network is trained, we train a linear classifier on top of this 10 dimensional output using Adam with identical configurations except a batch size of 128 is used. We early stop on the validation set of MNIST and report the test accuracy using that model. The classifier reaches $99.22\%$ test accuracy. This experiment shows that MNIST classes lie on nearly disjoint manifolds and that Neural Bayes-DML can correctly label them. As baseline, a linear classifier trained on features from a randomly initialized identical CNN architecture reaches $42.97\%$.

\end{document}